\documentclass[journal]{IEEEtran}
\usepackage{cite}

\ifCLASSINFOpdf
   \usepackage[pdftex]{graphicx}
\else
   \usepackage[dvips]{graphicx}
\fi
\usepackage{amsmath}
\usepackage{algorithmic}

\usepackage{array}

\ifCLASSOPTIONcompsoc
 \usepackage[caption=false,font=normalsize,labelfont=sf,textfont=sf]{subfig}
\else
 \usepackage[caption=false,font=footnotesize]{subfig}
\fi

\usepackage{stfloats}
\usepackage{url}

\usepackage{pgf}
\usepackage{xcolor}
\usepackage{physics}
\usepackage{notation}
\usepackage{multirow}
\usepackage{amssymb}
\usepackage{acronym}
\usepackage{psfrag}
\usepackage{soul}
\usepackage{tikz}
\usepackage[inline]{enumitem}
\usepackage[level=3]{LatexInclusion/messages}

\acrodef{pmf}[PMF]{probability mass function}
\acrodef{pdf}[PDF]{probability density function}
\acrodefplural{pdf}[PDFs]{probability density functions}
\acrodef{mmse}[MMSE]{minimum mean-square error}
\acrodef{cnn}[CNN]{convolutional neural network}
\acrodef{gnn}[GNN]{graph neural network}
\acrodefplural{gnn}[GNNs]{graph neural networks}
\acrodef{mpnn}[MPNN]{message passing neural network}
\acrodef{da}[DA]{data association}
\acrodef{bp}[BP]{belief propagation}
\acrodef{mlp}[MLP]{multi-layer perceptron}
\acrodef{fg}[FG]{Factor Graph}
\acrodef{nebp}[NEBP]{neural enhanced belief propagation}
\acrodef{roi}[ROI]{region of interest}
\acrodef{bev}[BEV]{bird's eye view}
\acrodef{mot}[MOT]{multi-object tracking}
\acrodef{iou}[IoU]{intersection over union}
\acrodef{giou}[GIoU]{generalized \ac{iou}}
\acrodef{po}[PO]{potential object}
\acrodef{ids}[IDS]{identity switches}
\acrodef{frag}[Frag]{fragments}
\acrodef{amota}[AMOTA]{average multi-object tracking accuracy}
\acrodef{amotp}[AMOTP]{average multi-object tracking precision}
\acrodef{nms}[NMS]{non-maximum suppression}
\acrodef{fp}[FP]{false positives}
\acrodef{mota}[MOTA]{multi-object tracking accuracy}
\acrodef{motp}[MOTP]{multi-object tracking precision}
\acrodef{jpda}[JPDA]{joint probabilistic data association}
\acrodef{mht}[MHT]{multiple hypothesis tracker}
\acrodef{rfs}[RFS]{random finite sets}
\acrodef{gospa}[GOSPA]{generalized optimal sub-pattern assignment}

\newcommand{\rd}{\textcolor{red}}
\newcommand{\ist}{\hspace*{.3mm}}
\newcommand{\rmv}{\hspace*{-.3mm}}

\newcommand{\nn}{\nonumber}
\newcommand{\T}{\mathrm{T}}
\newcommand{\acr}[1]{\aclu{#1} (\acs{#1})}

\definecolor{myblue}{RGB}{79, 129, 189}
\definecolor{myorange}{RGB}{247, 150, 70}

\begin{document}

\title{Neural Enhanced Belief Propagation \\ 
for Multiobject Tracking}

\author{Mingchao~Liang,~\IEEEmembership{Student Member,~IEEE,}
        and~Florian~Meyer,~\IEEEmembership{Member,~IEEE,} \vspace{-4mm}
        
                \thanks{This work was supported in part by the University of California San Diego and by the National Science Foundation (NSF) under CAREER Award No. 2146261. Parts of this work will be presented at the ISIF FUSION-22, Linköping, Sweden, July 2022.}

\thanks{Mingchao~Liang is with the Department of Electrical and Computer Engineering, University of California San Diego, La Jolla, CA, USA (e-mail: \texttt{m3liang@eng.ucsd.edu}).}

\thanks{Florian~Meyer is with the Scripps Institution of Oceanography and the Department of Electrical and Computer Engineering, University of California San Diego, La Jolla, CA, USA (e-mail: \texttt{flmeyer@ucsd.edu}).}
        
        }



\maketitle

\begin{abstract}
Algorithmic solutions for \ac{mot} are a key enabler for applications in autonomous navigation and applied ocean sciences. State-of-the-art \ac{mot} methods fully rely on a statistical model and typically use preprocessed sensor data as measurements. In particular, measurements are produced by a detector that extracts potential object locations from the raw sensor data collected for a discrete time step. This preparatory processing step reduces data flow and computational complexity but may result in a loss of information. State-of-the-art Bayesian \ac{mot} methods that are based on belief propagation \acused{bp}(\ac{bp}) systematically exploit graph structures of the statistical model to reduce computational complexity and improve scalability. However, as a fully model-based approach, BP can only provide suboptimal estimates when there is a mismatch between the statistical model and the true data-generating process. Existing BP-based \ac{mot} methods can further only make use of preprocessed measurements. In this paper, we introduce a variant of BP that combines model-based with data-driven \ac{mot}. The proposed \ac{nebp} method complements the statistical model of BP by information learned from raw sensor data. This approach conjectures that the learned information can reduce model mismatch and thus improve data association and false alarm rejection. Our NEBP method improves tracking performance compared to model-based methods. At the same time, it inherits the advantages of \ac{bp}-based \ac{mot}, i.e.,  it scales only quadratically in the number of objects, and it can thus generate and maintain a large number of object tracks.  We evaluate the performance of our \ac{nebp} approach for \ac{mot} on the nuScenes autonomous driving dataset and demonstrate that it has state-of-the-art performance.
\end{abstract}

\begin{IEEEkeywords}
Multiobject tracking, belief propagation, probabilistic data association, factor graphs, graph neural networks
\vspace{-2mm}
\end{IEEEkeywords}

%
\IEEEpeerreviewmaketitle

\section{Introduction}

\label{sec:introduction}

Multi-object tracking \acused{mot}(\ac{mot}) \cite{BarWilTia:B11,Mah:B07,MeyBraWilHla:J17,MeyKroWilLauHlaBraWin:J18,MeyWil:J21,Wil:J15,GarWilGraSve:J18,SchBenRosKriGra:18,SolMeyBraHla:J19,PanMorRad:21,LiuBaiXiaHuaZhu:22,ZhaMey:C21,YinZhoKra:21,MeyWin:J20,ChiPriLiBoh:20,WenWanHelKit:20,ZaeDaiLinDanVan:22,RanMahGebMhaRamTri:21,PanLiWan:21,WanChePanWanZha:21,WenWanManKit:20,ChiLiAmbBoh:21,LiLeiVen:J22} enables emerging applications including autonomous driving, applied ocean sciences, and indoor localization. \ac{mot} aims at estimating the states (e.g., positions and possibly other parameters) of moving objects over time, based on measurements provided by sensing technologies such as Light Detection and Ranging (LiDAR), radar, or sonar \cite{BarWilTia:B11,Mah:B07,MeyBraWilHla:J17,MeyKroWilLauHlaBraWin:J18,MeyWil:J21,Wil:J15,GarWilGraSve:J18,SchBenRosKriGra:18,SolMeyBraHla:J19,PanMorRad:21,LiuBaiXiaHuaZhu:22,ZhaMey:C21,YinZhoKra:21,MeyWin:J20,ChiPriLiBoh:20,WenWanHelKit:20,ZaeDaiLinDanVan:22,RanMahGebMhaRamTri:21,PanLiWan:21,WanChePanWanZha:21,WenWanManKit:20,ChiLiAmbBoh:21,LiLeiVen:J22}. An inherent problem in \ac{mot} is measurement-origin uncertainty, i.e., the unknown association between measurements and objects. \ac{mot} is further complicated by the fact that the number of objects is unknown, i.e., for the initialization and termination of object tracks, track management schemes need to be employed.

\subsection{Model-Based and Data-Driven MOT}

Typically \ac{mot} methods rely on measurements that have been extracted from raw sensor data in a detection process. For example, an object detector \cite{ZhuTuz:18, LanVorCaeZhoYanBei:19, ZhuJiaZhoLiYu:19, RenHeGirSun:15, ShiWanLi:19, SimBulPorLop:19} can be applied to LiDAR scans or images at each time step independently, and the detected objects are then used as measurements for \ac{mot} \cite{OkuTalFreLitLow:04, BreReiLeiKolVan:10}. This common strategy is referred to as ``detect-then-track''. Based on the assumption that, at each time step, an object can generate at most one measurement and each measurement can be originated by at most one object, data association can be cast as a bipartite matching problem.

The first class of \ac{mot} methods follows a global nearest neighbor association approach \cite{BarWilTia:B11}. Here, a Hungarian \cite{Kuh:55} or a greedy matching algorithm is used to perform ``hard'' measurement-to-object associations \cite{WenWanHelKit:20, WenWanManKit:20, ChiPriLiBoh:20, ChiLiAmbBoh:21, ZaeDaiLinDanVan:22, RanMahGebMhaRamTri:21, PanLiWan:21, WanChePanWanZha:21}. To improve the reliability of hard associations, these methods often rely on discriminative shape information of objects and measurements. Shape information is extracted from raw sensor data based on deep neural networks \cite{ChiLiAmbBoh:21, WenWanManKit:20} and used to compute pairwise distances between objects and measurements more accurately. The methods in this class typically rely on heuristics for track management.


A second class of methods formulates and solves the \ac{mot} problem in the Bayesian estimation framework \cite{BarWilTia:B11,Mah:B07,Wil:J15, MeyWil:J21,GarWilGraSve:J18, SchBenRosKriGra:18,PanMorRad:21,MeyBraWilHla:J17,MeyKroWilLauHlaBraWin:J18,ZhaMey:C21}. Methods in this class rely on a statistical model for object birth, object motion, and measurement generation \cite{BarWilTia:B11,Mah:B07,Wil:J15,MeyWil:J21,GarWilGraSve:J18, SchBenRosKriGra:18,PanMorRad:21,MeyBraWilHla:J17,MeyKroWilLauHlaBraWin:J18,ZhaMey:C21}. The statistical model makes it possible to perform a more robust probabilistic ``soft'' data association and to avoid heuristics for track initialization and termination. Methods in this class include the \ac{jpda} filter \cite{BarWilTia:B11}, the \ac{mht} \cite{Bla:J04}, \ac{rfs} filters \cite{Wil:J15,GarWilGraSve:J18,SchBenRosKriGra:18,PanMorRad:21, LiuBaiXiaHuaZhu:22}, and \ac{bp}-based \ac{mot} \cite{Wil:J15,MeyBraWilHla:J17,MeyKroWilLauHlaBraWin:J18,MeyWil:J21}.

\ac{bp}, also known as the sum-product algorithm, \cite{KscFreLoe:01,YedFreWei:05,KolFri:B09} provides an efficient and scalable solution to high-dimensional inference problems. \ac{bp} operates by ``passing messages'' along the edges of the factor graph \cite{KscFreLoe:01} that represents the statistical model of the inference problem. Important algorithms such as the Kalman filter, the particle filter \cite{AruMasGorCla:02}, and the \ac{jpda} filter \cite{BarWilTia:B11} are instances of \ac{bp}. By exploiting the structure of the graph, \ac{bp}-based \ac{mot} methods \cite{Wil:J15,MeyBraWilHla:J17,MeyKroWilLauHlaBraWin:J18,MeyWil:J21} are highly scalable. In particular, using BP, ``soft'' probabilistic data association can be performed for hundreds of objects. This makes it possible to generate and maintain a very large number of potential object tracks and, in turn, achieve state-of-the-art \ac{mot} performance \cite{Wil:J15,MeyBraWilHla:J17,MeyKroWilLauHlaBraWin:J18,MeyWil:J21}.  

Existing \ac{bp}-based methods entirely rely on ``hand-designed'' statistical models. However, the statistical models are often unable to accurately represent all the intricate details of the true data-generating process. This mismatch leads to suboptimal object state estimates. In addition, since \ac{bp} methods rely on the detect-then-track strategy, important object-related information might be discarded by the object detector. On the other hand, learning-based methods are fully data-driven, i.e., they do not make use of any statistical model.  Typically, learning-based methods rely on deep neural networks, which facilitate the extraction of all relevant information from raw sensor data \cite{WenWanManKit:20, ChiLiAmbBoh:21}. However, learning-based \ac{mot} typically makes use of potentially unreliable heuristics for track management and only performs well in ``big data'' problems.


A graph neural network (GNN) \cite{GorMonSca:05, GilSchRilVinDah:17} is a graphical model formed by neural networks. The neural network ``passes messages'', i.e., exchange processing results, along the edges of the GNN. This mechanism is similar to the message passing performed by BP. It has been demonstrated that in Bayesian estimation problems, a GNN can outperform loopy BP  \cite{YooLiaXioZhaFetUrtZemPit:19} if sufficient training data is available. Recently, neural enhanced belief propagation (NEBP)  \cite{SatWel:21} was introduced.  In \ac{nebp}, a \ac{gnn} \cite{GorMonSca:05, GilSchRilVinDah:17} that matches the topology of the factor graph is established. After training the GNN, the GNN messages can complement the corresponding BP messages to correct errors introduced by cycles and model mismatch. The resulting hybrid message passing method combines the benefits of model-based and data-driven inference. In particular, NEBP can leverage the performance advantages of GNNs in big data problems. The benefits of NEBP have been demonstrated in decoding \cite{SatWel:21} and cooperative localization \cite{LiaMey:21} problems.

\subsection{Contribution and Paper Organization}

In this paper, we address the fundamental question of how model-based and data-driven approaches can be combined in a hybrid inference method. In particular, we aim to develop a \ac{bp}-based \ac{mot} method that augments its ``hand-designed'' statistical model with information learned from raw sensor data.
As a result, we propose \ac{nebp} for \ac{mot}. Here, BP messages calculated for probabilistic data association are combined with the output of a \ac{gnn}. The \ac{gnn} uses object detections and features learned from raw sensor information as inputs. It can improve the MOT performance of BP by introducing data-driven \textit{false alarm rejection} and \textit{object shape association}. 

\textit{False alarm rejection} aims at identifying which measurements are likely false alarms. For measurements that have been identified as a potential false alarm, the false alarm distribution in the statistical model used by \ac{bp} is locally increased. This reduces the probability that the measurement is associated with an existing object track or initializes a new object track. \textit{Object shape association} computes improved association probabilities by also taking features of existing object tracks and measurements that have been learned from raw sensor data into account. Compared to \ac{bp} for \ac{mot}, the resulting \ac{nebp} method for \ac{mot} can improve object declaration and estimation performance if annotated data is available, and consequently provide state-of-the-art performance in big data \ac{mot} problems.

\begin{figure*}
    \centering
    \hspace{10mm}
    \psfrag{sm1}[c][c][0.8]{\rd{Shape and Motion}}
    \psfrag{fe1}[c][c][0.8]{\rd{Feature Extraction}}
    \psfrag{sf1}[l][l][0.8]{Shape Features $\V{h}_{\text{shape}}$}
    \psfrag{mf1}[l][l][0.8]{\hspace{-1.7mm}Motion Features $\V{h}_{\text{motion}}$}
    \psfrag{meas1}[c][c][0.8]{\hspace{2mm}Measurements $\V{z} = [\V{z}^{\T}_1 \dots \V{z}^{\T}_{J}]^{\T} $}
    \psfrag{gnn1}[c][c][0.8]{\rd{Graph Neural}}
    \psfrag{gnn2}[c][c][0.8]{\rd{Network (GNN)}}
    \psfrag{fg1}[c][c][0.8]{\raisebox{-4mm}{Factor Graph and BP}}
    \psfrag{gnn2}[c][c][0.8]{\rd{Network (GNN)}}
    \psfrag{objd1}[c][c][0.8]{Object}
    \psfrag{objd2}[c][c][0.8]{Detector}
    \psfrag{bp1}[r][r][0.8]{BP Messages $\V{\phi}$}
    \psfrag{nebp1}[l][l][0.8]{NEBP Messages $\tilde{\V{\phi}}$}
    \psfrag{rsd1}[r][r][0.8]{Raw Sensor Data $\Set{Z}$}
    \psfrag{rsd2}[r][r][0.8]{from Current Time}
    \psfrag{prsd1}[r][r][0.8]{Raw Sensor Data $\Set{Z}^{-}$}
    \psfrag{prsd2}[r][r][0.8]{from Previous Time}
    \psfrag{pbe1}[r][r][0.8]{Beliefs $\tilde{f}(\V{x}_{i}^{-}, r_{i}^{-}),$}
    \psfrag{pbe2}[r][r][0.8]{$i \rmv\in \{1, \dots, I^{-}\}$}
    \psfrag{pbe3}[r][r][0.8]{from Previous Time}
    \psfrag{pbe4}[r][r][0.8]{}
    \psfrag{be1}[r][r][0.8]{Beliefs $\tilde{f}(\V{x}_{i}, r_{i})$,}
    \psfrag{be2}[r][r][0.8]{$i \rmv\in \{1, \dots, I\}$}
    \psfrag{be3}[r][r][0.8]{from Current Time}
    \includegraphics[scale=0.6]{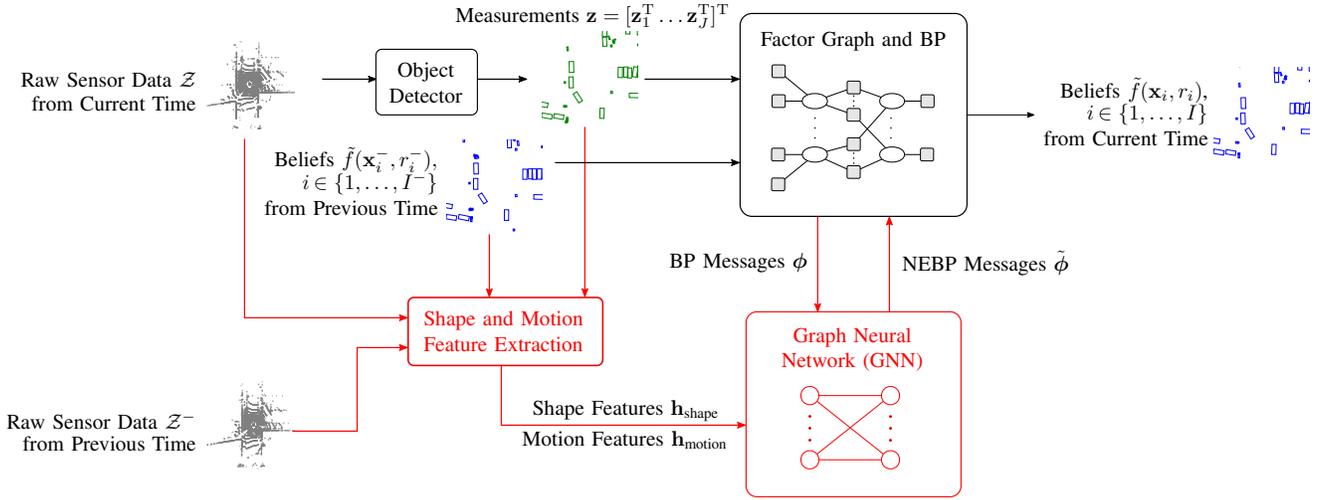}
    \caption{Flow diagram of one time step of conventional \ac{bp} and the proposed \ac{nebp} method for \ac{mot}. Black boxes show the computation modules performed by both \ac{bp} and \ac{nebp}. The red boxes show additional modules only performed by the proposed \ac{nebp} method. The goal of both \ac{mot} methods is to obtain estimates $\hat{\V{x}}_{i}$ and $\hat{r}_{i}$ of object state $\V{x}_{i}$ and existence variable $r_{i} \rmv\in\rmv \{0,1\}$ for all objects $i \in \{1,\dots,I\}$. First, raw sensor data $\Set{Z}$ is preprocessed by an object detector and the resulting object detection vector $\V{z}$ consists of the measurements used for \ac{mot}. In addition to the measurements $\V{z}$, approximate marginal posterior distributions (``beliefs'') $\tilde{f}\big(\V{x}_{i}^{-}\rmv, r_{i}^{-}\big)$, $i \rmv\in \{1, \dots, I^{-}\}$ that\vspace{-.0mm} have been computed at the previous time step are used as input for \ac{mot}. The \acr{bp}\vspace{-.0mm} method reviewed in Section \ref{subsec:fg_bp} performs operations on the factor graph discussed in Section \ref{sec:bp_mot} to compute updated beliefs $\tilde{f}(\V{x}_{i}, r_{i})$, $i \rmv\in \{1, \dots, I\}$.  Estimates $\hat{\V{x}}_{i}$ and $\hat{r}_{i}$ can be computed from the beliefs  $\tilde{f}(\V{x}_{i}, r_{i})$ as discussed in Section \ref{sec:estimationDetection}. Compared to conventional \ac{bp} for \ac{mot}, the proposed \ac{nebp} approach introduces a shape and motion feature extraction module and a \ac{gnn}, discussed in Section \ref{sec:feat_extract} and \ref{subsec:gnn_topo_nebp}, respectively. The shape and motion feature extraction module computes shape and motion features $\V{h}_{\text{shape}}$ and $\V{h}_{\text{motion}}$ from raw sensor data $\mathcal{Z}$ and $\mathcal{Z}^{-}$ of the previous and current time. The \ac{gnn} computes \ac{nebp} messages $\tilde{\V{\phi}}$ based on the conventional \ac{bp} messages $\V{\phi}$ and the features $\V{h}_{\text{shape}}$ and $\V{h}_{\text{motion}}$, to obtain more accurate beliefs $\tilde{f}(\V{x}_{i}, r_{i})$, $i \rmv\in \{1, \dots, I\}$ and thus more accurate estimates $\hat{\V{x}}_{i}$ and $\hat{r}_{i}$ as discussed in Section \ref{sec:statementMOT}.}
    \label{fig:overview}
\end{figure*}

The key contributions of this paper are summarized as follows.
\begin{itemize}
    \item We introduce \ac{nebp} for \ac{mot} where probabilistic data association is enhanced by learned information provided by a \ac{gnn}.
    \vspace{.8mm}
    
     \item We present the procedure and the loss function, used for the training of the \ac{gnn}, that enable false alarm rejection and object shape association.   
     \vspace{.8mm}
    
    \item We apply the proposed method to an autonomous driving dataset and demonstrate state-of-the-art object tracking performance.
 \end{itemize}
   An overview of the proposed \ac{nebp} method for \ac{mot} is presented as a flow diagram in Fig.~\ref{fig:overview}. Here, black boxes show the computation modules performed by both conventional \ac{bp} and \ac{nebp}. The red boxes show the additional modules only performed by the proposed \ac{nebp} method. A detailed description of each module will be provided in Sections \ref{sec:bp_mot} and \ref{sec:nebp_mot}.

In modern MOT scenarios with high-resolution sensors, it is often challenging to capture object shapes and the corresponding data-generating process by a statistical model. Thus, in contrast to the extended object tracking strategy \cite{GraBauReu:J17,GraFatSve:J19,MeyWil:J21}, the influence of object shapes on data generation is best learned directly from data. This paper advances over the preliminary account of our method provided in the conference publication \cite{LiaMey:C22} by (i) introducing a new factor graph representation, which is a more accurate description of the proposed \ac{nebp} method; (ii) presenting more details on the development and implementation of the proposed \ac{nebp} approach for \ac{mot}; and (iii) conducting a comprehensive evaluation based on real data that highlights why NEBP advances BP in MOT applications, and (iv) providing a detailed complexity analysis of the proposed \ac{nebp} for \ac{mot} method. Note that the new factor graph representation does not alter the resulting \ac{nebp} method for \ac{mot}.

This paper is organized as follows. Section \ref{sec:bp_nebp} reviews the general \ac{bp} and \ac{nebp} algorithm. Section \ref{sec:model} describes the system model and statistical formulation. Section \ref{sec:bp_mot} reviews the factor graph and the \ac{bp} for \ac{mot} algorithm. Section \ref{sec:nebp_mot} develops the proposed \ac{nebp} for \ac{mot} algorithm. Section \ref{sec:loss} introduces the loss function used for NEBP training. Section \ref{sec:exp} discusses experimental results and Section \ref{sec:conclusion} concludes the paper.

\section{Review of Belief Propagation and Neural Enhanced Belief Propagation} \label{sec:bp_nebp}

\subsection{Factor Graph and Belief Propagation} \label{subsec:fg_bp}
A factor graph \cite{KscFreLoe:01, Loe:04} is a bipartite undirected graph $\Set{G}_f \rmv=\rmv (\Set{V}_f, \Set{E}_f)$ that consists of a set of edges $\Set{E}_f$ and a set of vertices or nodes $\Set{V}_f = \Set{Q} \cup \Set{F}$. A variable node $q \in \Set{Q}$ represents a random variable $\V{x}_q$ and a factor node $s \in \Set{F}$ represents a factor $\psi_s\big(\V{x}^{(s)}\big)$. The argument $\V{x}^{(s)}$ of a factor, consists of certain random variables $\V{x}_q$ (each $\V{x}_q$ can appear in several $\V{x}^{(s)}$). Variable nodes and factor nodes are typically depicted by circles and boxes, respectively. The joint \ac{pdf} represented by the factor graph\vspace{2mm} reads 
\begin{equation}
    f(\V{x}) \propto \prod_{s \in \Set{F}} \psi_s\big(\V{x}^{(s)}\big)
\end{equation}
where $\propto$ denotes equality up to a multiplicative constant.

\ac{bp} \cite{KscFreLoe:01}, also known as the sum-product algorithm, can compute marginal \acp{pdf} $f(\V{x}_q)$, $q \rmv\in\rmv \Set{Q} $ efficiently. \ac{bp} performs local operations on the factor graph. The local operations can be interpreted as ``messages'' that are passed over the edges of the graph. There are two types of messages.
At message passing iteration $\ell \in \{1, \cdots, L\}$, the messages passed from variable nodes to factor nodes are defined\vspace{.5mm} as
\begin{equation}
    \phi^{(\ell)}_{q \to s}(\V{x}_q) = \prod_{a \in \Set{N}_{\Set{F}}(q) \backslash s} \phi^{(\ell-1)}_{a \to q}(\V{x}_q) \label{eq:bp_x_to_f}.
\end{equation}
In addition, the messages passed from factor nodes to variable nodes are given by
\begin{equation}
    \phi^{(\ell)}_{s \to q}(\V{x}_q) = \hspace{-0.7mm} \sum_{\V{x}^{(s)} \backslash \V{x}_q} \hspace{-0.7mm} \psi_s\big(\V{x}^{(s)}\big) \hspace{-0.7mm} \prod_{m \in \Set{N}_{\Set{Q}}(s) \backslash q} \phi^{(\ell)}_{m \to s}(\V{x}_m) \label{eq:bp_f_to_x}
\end{equation}
where $\Set{N}_{\Set{Q}}(\cdot) \rmv\subseteq\rmv \Set{Q}$ and $\Set{N}_{\Set{F}}(\cdot) \rmv\subseteq\rmv \Set{F}$ denote the set of neighboring variable and factor nodes, respectively. If $\psi_s\big(\V{x}^{(s)}\big)$ is a singleton factor node in the sense that it is connected to a single variable node $\V{x}_q$, i.e., $\V{x}^{(s)} = \V{x}_q$, then the message from the factor node to the variable node is equal to the factor node itself, i.e., $\phi_{q}(\V{x}_q) \triangleq \phi_{s \to q}(\V{x}_q) \rmv=\rmv \psi_s\big(\V{x}_q\big) $. For future use, we introduce the joint set of messages $\V{\phi}^{(\ell + 1)} = \big\{\phi^{(\ell + 1)}_{q}, \phi^{(\ell + 1)}_{q \to s},\phi^{(\ell + 1)}_{s \to q}\big\}_{q \in \Set{Q}, s \in \Set{F}}$ and\vspace{-.8mm} $\V{\phi}_{\Set{F} \to \Set{Q}}^{(\ell + 1)} = \big\{\phi^{(\ell + 1)}_{q}\rmv\rmv,\phi^{(\ell + 1)}_{s \to q} \big\}_{q \in \Set{Q}, s \in \Set{F}}$ as well as\vspace{.5mm} the function
\begin{equation}
\V{\phi}^{(\ell)} \rmv\rmv = \text{BP}\Big(\V{\phi}_{\Set{F} \to \Set{Q}}^{(\ell-1)}\Big) \nn
\vspace{.5mm}
\end{equation}
that summarizes all message computations \eqref{eq:bp_x_to_f} and \eqref{eq:bp_f_to_x} related to one iteration $\ell$.

After message passing is completed, one can subsequently obtain ``beliefs'' $\tilde{f}(\V{x}_q)$, for each variable node $\V{x}_q$, computed as the product of all incoming messages\vspace{1mm}, i.e.,
\begin{equation}
    \tilde{f}(\V{x}_q) \propto \prod_{s \in \Set{N}_{\Set{F}}(q)} \phi^{(L)}_{s \to q}(\V{x}_q).
\end{equation} 

If the factor graph is a tree, then the beliefs are exactly equal to the marginal \ac{pdf}, i.e., $\tilde{f}(\V{x}_q) = f(\V{x}_q)$. In factor graphs with loops, \ac{bp} is applied in an iterative manner and the message passing order is not unique. Different message passing orders may lead to different beliefs. The beliefs $\tilde{f}(\V{x}_q)$ provided by this ``loopy \ac{bp}'' scheme are only approximations of marginal posterior \acp{pdf} $f(\V{x}_q)$. However, the beliefs $\tilde{f}(\V{x}_q)$ have been observed to be very accurate in many applications\cite{Wym:07,WaiJor:03,MeyKroWilLauHlaBraWin:J18}.
\vspace{-2.5mm}

\subsection{Neural Enhanced Belief Propagation} \label{subsec:nebp}

\ac{nebp} \cite{SatWel:21} is a hybrid message passing method that combines the benefits of model-based and data-driven inference. In particular, \ac{nebp} aims at improving the \ac{bp} solution by augmenting the factor graph by a \ac{gnn}. While \ac{bp} messages are calculated based on the statistical model represented by the factor graph, \ac{gnn} messages are computed based on information learned from data. In \ac{nebp}, a \ac{gnn} that matches the network topology of the factor graph is introduced. An iterative message passing procedure is performed on the \ac{gnn}. In one \ac{gnn} iteration, nodes send messages to their neighboring nodes (cf. \eqref{eq:nebp_x_to_f}-\eqref{eq:nebp_f_to_x}), receive messages from neighboring nodes, and aggregate received messages to update their node embeddings (cf. \eqref{eq:nebp_f}-\eqref{eq:nebp_x}). In particular, at message passing iteration $\ell \in \{1, \cdots, L\}$, the equations that describe message passing of the \ac{gnn}\vspace{.5mm} are given as follows \cite{SatWel:21}. The messages exchanged between variable nodes $q \in \Set{Q}$ to factor nodes $s \in \Set{F}$ along the edges of the \ac{gnn} are given by the vectors\vspace{.4mm}
\begin{align}
    \V{m}^{(\ell+1)}_{q \to s} &= g_{\Set{Q} \to \Set{F}}\big(\V{h}^{(\ell)}_{q}, \V{h}^{(\ell)}_{s}, \V{e}_{q \to s}\big) \label{eq:nebp_x_to_f}     
    \end{align}
    \vspace{-4mm}
    \begin{align}
    \V{m}^{(\ell+1)}_{s \to q} &= g_{\Set{F} \to \Set{Q}}\big(\V{h}^{(\ell)}_{s}, \V{h}^{(\ell)}_{q}, \V{e}_{s \to q}\big) \label{eq:nebp_f_to_x} 
\end{align}
where the $\V{e}_{q \to s}$ as well as the $\V{e}_{s \to q}$, are edge attribute vectors and $g_{\Set{Q} \to \Set{F}}(\cdot)$ as well as  $g_{\Set{F} \to \Set{Q}}(\cdot)$ are referred to as edge functions \cite{SatWel:21}. In addition, after GNN messages have been exchanged, so-called node embedding vectors $\V{h}^{(\ell)}_{s}$ and $\V{h}^{(\ell)}_{q}$ for factor node $s \in \Set{F}$ and variable node $q \in \Set{Q}$, are computed as
\begin{align}
    \V{h}^{(\ell + 1)}_{s} &= g_{\Set{F}}\Big(\V{h}^{(\ell)}_{s}, \sum_{q \in \Set{N}_{\Set{Q}}(s)} \V{m}^{(\ell+1)}_{q \to s}\Big) \label{eq:nebp_f} \\[1mm]
    \V{h}^{(\ell + 1)}_{q} &= g_{\Set{Q}}\Big(\V{h}^{(\ell)}_{q}, \sum_{s \in \Set{N}_{\Set{F}}(q)} \V{m}^{(\ell+1)}_{s \to q}, \V{e}_{q}\Big)\label{eq:nebp_x} \\[-5.7mm]
    \nn
\end{align}
where the $\V{e}_{q}$ are node attribute vectors \cite{SatWel:21} and $g_{\Set{F}}(\cdot)$ as well as $g_{\Set{Q}}(\cdot)$ are referred to as node functions \cite{SatWel:21}. Edge and node functions are the neural networks of the \ac{gnn}. For future use, we introduce the joint set of GNN messages $\V{m}^{(\ell)} = \{\V{m}^{(\ell)}_{s \to q}, \V{m}^{(\ell)}_{q \to s}\}_{q \in \Set{Q}, s \in \Set{F}}$, node embeddings $\V{h}^{(\ell)} = \{\V{h}^{(\ell)}_{s}\rmv\rmv, \V{h}^{(\ell)}_{q} \}_{q \in \Set{Q}, s \in \Set{F}}$, and attributes $\V{e} = \{\V{e}_{q}, \V{e}_{s\to q}, \V{e}_{q\to s}\}_{q \in \Set{Q}, s \in \Set{F}}$  as well as\vspace{.3mm} the function $\big\{\V{h}^{(\ell + 1)}\rmv\rmv, \V{m}^{(\ell + 1)} \big\}=\text{GNN}(\V{h}^{(\ell)}\rmv\rmv, \V{e})$ that summarizes all GNN computations \eqref{eq:nebp_x_to_f}--\eqref{eq:nebp_x} at iteration $\ell$. Note that singleton factor nodes are not included in the GNN.

Based on the BP message passing procedure discussed in Section \ref{subsec:fg_bp} and the GNN message passing procedure discussed above, the hybrid \ac{nebp} method can be summarized as follows. In particular, at iteration $\ell$ 
\begin{align}
    \V{\phi}^{(\ell + 1)} &= \text{BP}\Big(\tilde{\V{\phi}}_{\Set{F} \to \Set{Q}}^{(\ell)}\Big) \nn \\[2mm]
    \big\{\V{h}^{(\ell + 1)}\rmv\rmv, \V{m}^{(\ell + 1)} \big\}&= \text{GNN}\Big(\V{h}^{(\ell)}\rmv\rmv, \V{\phi}^{(\ell + 1)}\rmv\Big) \nn
\end{align}
where $\tilde{\V{\phi}}_{\Set{F} \to \Set{Q}}^{(\ell)} = \big\{\tilde{\phi}^{(\ell)}_{s \to q} \big\}_{q \in \Set{Q}, s \in \Set{F}}$ is\vspace{.4mm} the set of \ac{nebp} messages from the last iteration $\ell$ that are passed from factor nodes to variables nodes.
The \ac{bp} messages $\V{\phi}^{(\ell + 1)}$ serve as the edge attributes for \ac{gnn} message passing in \eqref{eq:nebp_x_to_f}-\eqref{eq:nebp_x}. This can be seen as providing a preliminary data association solution computed by conventional \ac{bp}, which does not make use of the object shape information, to the \ac{gnn}. The GNN then aims at refining this preliminary solution by also taking the object shape information into account. Providing a preliminary solution to the \ac{gnn} can make training and inference more efficient and accurate \cite{SatWel:21}. 

Finally, the \ac{nebp} messages at the current iteration, are calculated\vspace{1mm} as
\begin{align}
    \tilde{\phi}^{(\ell + 1)}_{s \to q} &= g_{\text{nebp},1}\Big(   \V{h}^{(\ell+1)}\rmv\rmv\rmv, \V{m}^{(\ell+1)}   \Big) \circ \phi^{(\ell + 1)}_{s \to q} \nn\\[.5mm]
    &\hspace{0mm}+ g_{\text{nebp},2}\Big(\V{h}^{(\ell+1)}\rmv\rmv\rmv, \V{m}^{(\ell+1)}\Big) \label{eq:nebp_ori_comb} \\[-6mm]
\nn
\end{align}
where $g_{\text{nebp},1}\big(   \V{h}^{(\ell)}\rmv\rmv\rmv, \V{m}^{(\ell)}   \big) $ and $g_{\text{nebp},2}\big(\V{h}^{(\ell)}\rmv\rmv\rmv, \V{m}^{(\ell)}\big)$ are neural networks that, in general, output a positive vector with the same dimension as $\phi^{(\ell + 1)}_{s \to q}$ and $\circ$ is element-wise multiplication. The \ac{bp} messages passed from variable nodes to factor nodes are not neural enhanced.

After the last message passing iteration $\ell = L$, the beliefs for each variable node $\V{x}_q$ are calculated as\vspace{1mm}
\begin{equation}
\tilde{f}(\V{x}_{q}) = \prod_{s \in \Set{N}_{\Set{F}}(q)} \tilde{\phi}^{(L)}_{s \to q}(\V{x}_q).
\vspace{-1mm}
\end{equation}



\section{System Model and Statistical Formulation} \label{sec:model}

In this section, we review the system model of BP-based MOT and the multiobject declaration and state estimation problem BP-based MOT aims to solve.

\subsection{Potential Objects and Object States}
\label{sec:potObjects}
The number of objects is unknown and time-varying. We describe this scenario by introducing $N_k$ \acp{po} \cite{MeyBraWilHla:J17,MeyKroWilLauHlaBraWin:J18} where $N_k$  is the maximum possible number of objects\footnote{The number of POs $N_k$ is the maximum possible number of actual objects that have produced a measurement so far \cite{MeyKroWilLauHlaBraWin:J18}.}. At time $k$, the existence of a \ac{po} $n \in \{1, \dots, N_k\}$ is modeled by a binary random variable $r_{k, n} \in \{0, 1\}$. \ac{po} $n$ exists, in the sense that it represents an actual object, if and only if $r_{k, n} = 1$. The kinematic state of \ac{po} $n$ is modeled by a random vector $\V{x}_{k, n}$ that consists of the object's position and possibly motion information. The augmented \ac{po} state is defined as $\V{y}_{k, n} \triangleq [\V{x}_{k, n}^\T \hspace{1mm} r_{k, n}]^\T$. In what follows, we will refer to augmented \ac{po} states simply as PO states. We also introduce the joint \ac{po} state vector at time $k$ as $\V{y}_{k} \triangleq [\V{y}_{k, 1}^\T \cdots \V{y}_{k, N_k}^\T]^\T$.

At time $k$, a detector $g_{\mathrm{det}}(\cdot)$ produces a vector $\V{z}_{k} \triangleq [\V{z}_{k, 1}^\T \cdots \V{z}_{k, J_k}^\T]^\T$ of preprocessed measurements from raw sensor data $\Set{Z}_k$, i.e., $\V{z}_{k}  = g_{\mathrm{det}}(\Set{Z}_k)$. The joint measurement vector that consists of all preprocessed measurements up to time $k$ is denoted as $\V{z}_{1 : k} \triangleq [\V{z}_{1}^\T \cdots \V{z}_{k}^\T]^\T$.

There are two types of \acp{po}:
\begin{itemize}
    \item \textit{New \acp{po}} represent objects that, for the first time, have generated a measurement at the current time step $k$. Their states are denoted as $\overline{\V{y}}{}_{k, j} = [\overline{\V{x}}{}_{k, j}^\T \hspace{1mm} \overline{r}{}_{k, j}]^\T$. At time $k$, a new \ac{po} is introduced\footnote{Introducing a new \ac{po} is equal to initializing a new potential object track \cite{MeyKroWilLauHlaBraWin:J18}.} for each measurement $\V{z}_{k, j}, j \in \{1, \dots, J_k\}$.
    \vspace{.8mm}
    
    \item \textit{Legacy \acp{po}} represent objects that already have generated at least one measurement at previous time steps $k^\prime < k$. Their states are denoted by $\underline{\V{y}}{}_{k, i} = [\underline{\V{x}}{}_{k, i}^\T \hspace{1mm} \underline{r}{}_{k, i}]^\T, i = \{1, \dots, I_k \}$, where $I_k$ is the total \vspace{.2mm} number of legacy \acp{po}.
\end{itemize}

All new \acp{po} that have been introduced at time $k-1$ become legacy \acp{po} at time $k$. Thus, the number of legacy \acp{po} at time $k$ is $I_k = I_{k - 1} + J_{k - 1} = N_{k - 1}$ and the total number of \acp{po} is $N_k = I_k + J_k$. A pruning step that limits the growth of the number of \ac{po} states will be discussed in Section~\ref{sec:estimationDetection}. For future reference, we further define the joint \ac{po} states $\overline{\V{y}}{}_{k} \triangleq [\overline{\V{y}}{}_{k, 1}^\T \cdots \overline{\V{y}}{}_{k, J_k}^\T ]^\T\rmv\rmv$, $\underline{\V{y}}{}_{k} \triangleq [\underline{\V{y}}{}_{k, 1}^\T \cdots \underline{\V{y}}{}_{k, I_k}^\T ]^\T\rmv\rmv$, and $\V{y}_k \triangleq [\underline{\V{y}}{}_{k}^\T \hspace{1mm} \overline{\V{y}}{}_{k}^\T]^\T\rmv\rmv$. 

\acp{po} represent actual objects that already have generated at least one measurement. In addition, there may also be actual objects that have not generated any measurements yet. These objects are referred to as ``unknown'' objects. Unknown objects are independent and identically distributed according to $f_{\text{u}}(\cdot)$. The number of unknown objects is modeled by a Poisson distribution with mean $\mu_{\text{u}}$. The statistical model for unknown objects induces a statistical model for new \acp{po} \cite{MeyKroWilLauHlaBraWin:J18} as further discussed in Section~\ref{sec:bp_mot}.



\subsection{Data Association Vector and Measurement Model} 
\label{subsec:model_meas}
\ac{mot} is subject to measurement origin uncertainty, i.e., it is unknown which actual object generates which measurement $\V{z}_{k, j}$. It is also possible that a measurement is not originated from any actual object. Such a measurement is referred to as a false alarm. Furthermore, an actual object may also not generate any measurements. This is referred to as missed detection. We assume that an object can generate at most one measurement and a measurement can originate from at most one object; this is known as the ``data association assumption.''

Since every actual object that has generated a measurement is represented by a PO, we can model measurement origin uncertainty by \ac{po}-to-measurement associations. These associations are represented by multinoulli random variables. In particular, the \ac{po}-to-measurement association at time $k$ can be described by the ``object-oriented'' \ac{da} vector $\V{a}_k = [a_{k, 1} \cdots a_{k, I_k}]^\T\rmv$.  The case where legacy \ac{po} $i$ generates measurement $j$ at time $k$, is represented by $a_{k, i} = j \in \{1, \dots, J_k\}$. On the other hand, the case where legacy \ac{po} $i$ does not generate any measurement at time $k$ is represented by $a_{k, i} = 0$. 

The computation complexity of \ac{mot} can be reduced by also introducing the ``measurement-oriented'' DA vector \cite{BaySahSha:J08,WilLau:10} $\V{b}_k = [b_{k, 1} \cdots b_{k, J_k}]^\T\rmv$. Here, $b_{k, j} = i \in \{1, \cdots, I_k\}$ represents the case where measurement $j$ is originated by legacy \ac{po} $i$. In addition, $b_{k, j} = 0$ represents the case where measurement $j$ is not originated by any legacy \ac{po}.  Modeling \ac{po}-to-measurement associations in terms of both $\V{a}_k$ and $\V{b}_k$ is redundant in that $\V{b}_k$ can be determined from $\V{a}_k$ and vice versa. However, the resulting hybrid representation makes it possible to check the consistency of the data association assumption based on indicators that are only a function of two scalar association variables. In particular, we introduce the indicator function $\Psi_{i, j}(a_{k, i}, b_{k, j})$ \cite{BaySahSha:J08,WilLau:10} which is equal to $0$ if $a_{k, i} = j, b_{k, j} \ne i$ or $a_{k, i} \ne j, b_{k, j} = i$, and is equal to $1$ otherwise.
 If and only if a data association event can be expressed by both an object-oriented $\V{a}_k$ and a measurement-oriented association vector $\V{b}_k$, then the event does not violate the data association assumption and all indicator functions are equal to one. Finally, we also introduce the joint \ac{da} vectors $\V{a}_{1 : k} \triangleq [\V{a}_1^\T \cdots \V{a}_k^\T]^\T$ and $\V{b}_{1 : k} \triangleq [\V{b}_1^\T \cdots \V{b}_{k}^\T]^\T$\rmv\rmv. It is assumed that an actual object generates a measurement with probability of detection $p_{\text{d}}$. If and only if a PO $i$ represents an actual object, i.e., $r_{k, i} \rmv=\rmv 1$, it can generate a measurement. If measurement $\V{z}_{k, j}$ has been generated by (legacy or new) \ac{po} $n \in \{1, \dots, N_k\}$, its conditional \ac{pdf} given \ac{po} state $\V{x}_{k, n}$ is  $f(\V{z}_{k, j} | \V{x}_{k, n})$. The functional form of $f(\V{z}_{k, j} | \V{x}_{k, n})$ is arbitrary. For example, if we have a linear measurement with respect to \ac{po} state $\V{x}_{k, n}$ with zero-mean, additive Gaussian noise, i.e. $\V{z}_{k, j} = \M{H}_k\V{x}_{k, n} \rmv+\rmv \V{v}_{k, j} $, then $f(\V{z}_{k, j} | \V{x}_{k, n})$ is given by $\Set{N}(\M{H}_k\V{x}_{k, n}, \M{R}_k)$, where $\M{R}_k$ is the covariance of the measurement noise $\V{v}_{k, j}$.

 If measurement $\V{z}_{k, j}$ has not been generated any \ac{po}, it is a false alarm measurement. False alarm measurements are independent and identically distributed according to $f_{\text{FA}} (\V{z}_{k, j})$. The number of false alarm measurements is modeled by a Poisson distribution with mean $\mu_{\text{FA}}$.

\subsection{Object Dynamics} 
The \ac{po} states $\V{y}_{k - 1, i}$ are assumed to evolve independently and identically according to a Markovian dynamic model \cite{BarWilTia:B11}. In addition, for each \ac{po} at time $k - 1$, there is a legacy \ac{po} at time $k$.  The state transition function of the joint \ac{po} state $\V{y}_{k - 1}$ at time $k - 1$, can thus be expressed as
\begin{equation}
    f(\underline{\V{y}}{}_{k} | \V{y}_{k - 1}) = \prod_{i = 1}^{N_{k - 1}} f(\underline{\V{y}}{}_{k, i} | \V{y}_{k - 1, i}) \label{eq:dynamics}
\end{equation}
where the state-transition \ac{pdf} $f(\underline{\V{y}}{}_{k, i} | \V{y}_{k - 1, i}) = f(\underline{\V{x}}{}_{k, i}, \underline{r}{}_{k, i} |$ $\V{x}_{k - 1, i}, r_{k - 1, i})$ models the dynamics of individual \acp{po} and is given as follows. If \ac{po} $i$ does not exist at time $k - 1$, i.e., $r_{k - 1, i} = 0$, then it cannot exist at time $k$ either. The state-transition \ac{pdf} for $r_{k - 1, i} = 0$ is thus given by
\begin{equation}
    f(\underline{\V{x}}{}_{k, i}, \underline{r}{}_{k, i} | \V{x}_{k - 1, i}, 0) \\ 
    = \begin{cases}
    f_{\text{D}}(\underline{\V{x}}{}_{k, i}), & \underline{r}{}_{k, i} = 0 \\
    0, & \underline{r}{}_{k, i} = 1
    \end{cases} \nn
\end{equation}
where $f_{\text{D}}(\underline{\V{x}}{}_{k, i})$ is an arbitrary ``dummy'' \ac{pdf} since states of nonexisting \acp{po} are irrelevant. If \ac{po} $i$ exists at time $k - 1$, i.e., $r_{k - 1, i} = 1$, 
then, the probability that it stills exists at time $k$ is given by the survival probability $p_{\mathrm{s}}$. If \ac{po} $i$ still exists at time $k$, its state $\underline{\V{x}}{}_{k, i}$ is distributed according to $f(\underline{\V{x}}{}_{k, i} | \V{x}_{k - 1, i})$. The state-transition \ac{pdf} for $r_{k - 1, i} = 1$, is thus given by
\begin{equation}
    f(\underline{\V{x}}{}_{k, i}, \underline{r}{}_{k, i} | \V{x}_{k - 1, i}, 1) \\ 
    = \begin{cases}
    (1 - p_{\mathrm{s}}) f_{\text{D}}(\underline{\V{x}}{}_{k, i}), & \underline{r}{}_{k, i} = 0 \\
    p_{\mathrm{s}} f(\underline{\V{x}}{}_{k, i} | \V{x}_{k - 1, i}), & \underline{r}{}_{k, i} = 1.
    \end{cases} \nn
    \vspace{-2mm}
\end{equation}
 
 
 
\subsection{Declaration, Estimation, Initialization, and Termination}  
\label{sec:estimationDetection}

At each time step $k$, our goal is to declare whether a \ac{po} $n \in \{1, \dots, N_k\}$ exists and to estimate the \ac{po} states $\V{x}_{k, n}$ of all existing \acp{po}, based on all measurements $\V{z}_{1 : k} = [\V{z}_{1}^\T \cdots \V{z}_{k}^\T]^\T\rmv$. In the Bayesian setting, object declaration and state estimation essentially amount to, respectively, calculating the marginal posterior existence probabilities $p(r_{k, n} \rmv=\rmv 1| \V{z}_{1: k})$ and the marginal posterior state \acp{pdf} $f(\V{x}_{k, n} | r_{k, n} = 1, \V{z}_{1: k})$. Then, a \ac{po} $n$ is declared to exist if $p(r_{k, n} = 1| \V{z}_{1  : k})$ is larger than a suitably chosen threshold $T_{\text{dec}}$ \cite[Ch. 2]{Poo:B94}. Furthermore, for each declared \ac{po} $n$, an estimate of $\V{x}_{k, n}$ is provided by the \ac{mmse} estimator \cite[Ch. 4]{Poo:B94}
\begin{equation}
    \hat{\V{x}}{}_{k, n} = \int \V{x}_{k, n} f(\V{x}_{k, n} | r_{k, n} = 1, \V{z}_{1 : k}) \mathrm{d}\V{x}_{k, n}.
\end{equation}
Both $p(r_{k, n} \rmv=\rmv 1| \V{z}_{1  : k})$ and $f(\V{x}_{k, n} | r_{k, n} \rmv=\rmv 1, \V{z}_{1 : k})$ can be obtained from the marginal posterior \acp{pdf} of augmented states $f(\V{y}_{k, n} | \V{z}_{1 : k}) = f(\V{x}_{k, n} , r_{k, n} | \V{z}_{1 : k})$. Thus, the problem to be solved is finding an efficient computation of $f(\V{y}_{k, n} | \V{z}_{1: k})$. For future reference, we introduce the notation $\hat{r}_{k, n} \rmv=\rmv p(r_{k, n} \rmv=\rmv 1| \V{z}_{1  : k})$.


Track initialization and termination can be summarized as follows. We initialize a new potential object track for each measurement. The initial existence probability of each potential object track is determined by the statistical model for unknown objects discussed in Section \ref{subsec:model_meas}. With this initialization approach, the number of object tracks grows linearly with time $k$. Therefore, we terminate (``prune'') potential object tracks by removing legacy and new \acp{po} with existence probabilities below a threshold $T_{\text{pru}}$ from the state space.

 





\begin{figure*}
    \centering
    \subfloat[]{
        \psfrag{g11}[c][c][0.7]{{$\Psi_{1,1}$}}
        \psfrag{g1I}[c][c][0.7]{$\Psi_{1,J}$}
        \psfrag{gJ1}[c][c][0.7]{$\Psi_{I,1}$}
        \psfrag{gIJ}[c][c][0.7]{$\Psi_{I,J}$}
    	\psfrag{ma1}[c][c][0.8]{\color{blue}{$q_{1}$}}
    	\psfrag{mb1}[c][c][0.8]{\color{blue}{$v_{1}$}}
    	\psfrag{mab11}[c][c][0.8]{\color{blue}{$\varphi_{1, 1}$}}
    	\psfrag{mbaJ1}[c][c][0.8]{\color{blue}{$\nu_{J, 1}$}}
    	\psfrag{mag11}[c][c][0.8]{\raisebox{-5mm}{\color{blue}{$\beta_{1, 1}$}}}
    	\psfrag{mbgJ1}[l][l][0.8]{\color{blue}{$\xi_{J, 1}$}}
    	\psfrag{q1}[c][c][0.8]{$q_1$}
    	\psfrag{qI}[c][c][0.8]{$q_I$}
    	\psfrag{v1}[c][c][0.8]{$v_1$}
    	\psfrag{vJ}[c][c][0.8]{$v_J$}
    	\psfrag{ya1}[c][c][0.8]{$\underline{\V{y}}{}_1, a_{1}$}
    	\psfrag{yaI}[c][c][0.8]{$\underline{\V{y}}{}_I, a_{I}$}
    	\psfrag{yb1}[c][c][0.8]{$\overline{\V{y}}{}_1, b_{1}$}
    	\psfrag{ybJ}[c][c][0.8]{$\overline{\V{y}}{}_J, b_{J}$}
    	\psfrag{p1}[c][c][0.8]{\color{blue}{$\alpha_1$}}
    	\psfrag{f1}[c][c][0.8]{$f_1$}
    	\psfrag{fI}[c][c][0.8]{$f_I$} \hspace{-10mm}
        \includegraphics[scale=0.8]{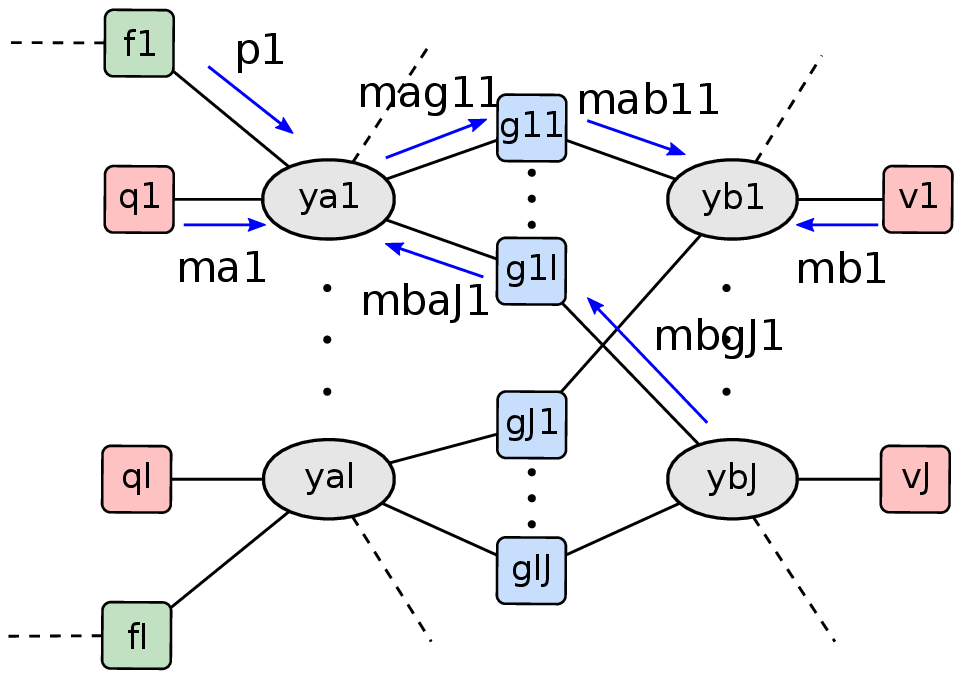}
        \label{fig:fg}}
    \hspace{13mm}
    \subfloat[]{\raisebox{12mm}{
        \psfrag{da1}[c][c][0.8]{$\V{h}_{a_{1}}$}
        \psfrag{dan}[c][c][0.8]{$\V{h}_{a_{I}}$}
        \psfrag{db1}[c][c][0.8]{$\V{h}_{b_{1}}$}
        \psfrag{dbm}[c][c][0.8]{$\V{h}_{b_{J}}$}
    	\psfrag{mab11}[c][c][0.8]{\color{red}{$\V{m}_{a_1 \to b_1}$}}
        \includegraphics[scale=0.8]{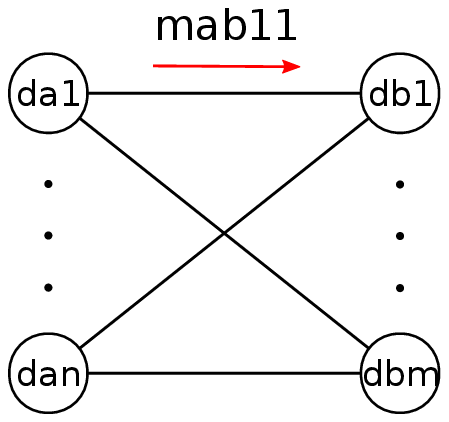}} 
        \label{fig:gnn}}
    
    \captionsetup{singlelinecheck = false, justification=justified}	
    \caption{Factor graph (a) and \vspace{-.5mm} corresponding bipartite \acr{gnn} (b) for a\vspace{.7mm} single time step $k$ of the considered NEBP approach for MOT. \ac{bp} and \ac{gnn} messages are shown. The time index $k$ is omitted. A GNN node was introduced for each legacy PO and each new PO. The topology of the \ac{gnn}, which only matches the part of the factor graph that models the data generating process, will be discussed in Section \ref{subsec:gnn_topo_nebp}. Following the topology of the data association part of the factor graph in (a), GNN edges were introduced such that the bipartite GNN shown in (b) is obtained. The following shorthand notation is used: $\underline{\V{y}}{}_i = \underline{\V{y}}{}_{k, i}$ $\overline{\V{y}}{}_j = \overline{\V{y}}{}_{k, j}$, $a_i = a_{k, i}$, $b_j = b_{k, j}$, $\Psi_{i, j} = \Psi_{i, j}(a_{k, i}, b_{k, j})$, $q_i = q(\underline{\V{x}}{}_{k, i}, \underline{r}{}_{k, i}, a_{k, i}; \V{z}_{k})$, $v_j = v(\overline{\V{x}}{}_{k, j}, \overline{r}{}_{k, j}, b_{k, j}; \V{z}_{k, j})$, $f_i = f(\underline{\V{y}}{}_{k, i} | \V{y}_{k - 1, i}), \alpha_{i} = \alpha_{k,i}(\underline{\V{x}}{}_{k, i}, \underline{r}{}_{k, i}), \beta_{i, j} = \beta^{(\ell)}_{k,i,j}(\underline{\V{y}}{}_{k, i},a_{k, i}), \xi_{j, i} = \xi^{(\ell)}_{k, j,i}(\overline{\V{y}}{}_{k, j},b_{k, j})$.}
    \label{fig:fg_gnn}
    \vspace{-3mm}
\end{figure*}






\section{Conventional BP-based MOT Algorithm} \label{sec:bp_mot}
In this section, we review the \ac{bp}-based \ac{mot} approach. Contrary to the original \ac{bp}-based \ac{mot} approach, we introduce an alternative factor graph which makes it easier to describe the proposed \ac{nebp} method presented in Section \vspace{0mm}\ref{sec:nebp_mot}.



By using common assumptions, the factorization structure of the joint posterior \ac{pdf} $f(\V{y}_{0: k}, \V{a}_{1: k}, \V{b}_{1: k} | \V{z}_{1: k})$ is given as follows (cf. \cite[Sec. VIII-G]{MeyKroWilLauHlaBraWin:J18}) \vspace{0mm}

\begin{alignat}{2}
    &f(\V{y}_{0 : k}, \V{a}_{1 : k}, \V{b}_{1 : k} | \V{z}_{1 : k})  \nn \\[1mm]
    & \propto \bigg( \prod_{n' = 1}^{N_0} f(\V{y}_{0, n'} ) \bigg) \prod_{k^\prime = 1}^{k} \bigg( \prod_{n = 1}^{N_{k^\prime - 1}} f(\underline{\V{y}}{}_{k^\prime\rmv, n} | \V{y}_{k^\prime - 1, n}) \bigg) \nn \\[1mm]
    & \hspace{.3mm}\times \bigg( \ist\ist\ist \prod_{i = 1}^{I_{k^\prime}} q(\underline{\V{y}}{}_{k^\prime\rmv, i}, a_{k^\prime\rmv, i}; \V{z}_{k^\prime})  \prod_{j^\prime = 1}^{J_{k^\prime}} \Psi_{i, j^\prime} (a_{k^\prime\rmv, i}, b_{k^\prime\rmv, j^\prime}) \bigg) \nn \\
    & \times \hspace{4.3mm}\prod_{j = 1}^{J_{k^\prime}} v(\overline{\V{y}}{}_{k^\prime\rmv, j}, b_{k^\prime\rmv, j}; \V{z}_{k^\prime, j}).
    \label{eq:factorization}
\end{alignat}
Note that often there are no \acp{po} at time $k = 0$, i.e., $N_0 \rmv=\rmv 0$. 

The factor $q(\underline{\V{y}}{}_{k, i}, a_{k, i}; \V{z}_{k}) \rmv\triangleq\rmv q(\underline{\V{x}}{}_{k, i}, \underline{r}{}_{k, i}, a_{k, i}; \V{z}_{k})$, describing the measurement model of the sensor for legacy \acp{po}, is defined\vspace{-2mm} as
\begin{align}
    q(\underline{\V{x}}{}_{k, i}, 1, a_{k, i}; \V{z}_{k}) &\triangleq 
    \begin{cases}
        \frac{\displaystyle p_{\text{d}} f(\V{z}_{k, j} | \underline{\V{x}}{}_{k, i}) }{\displaystyle \mu_{\text{FA}} f_{\text{FA}} (\V{z}_{k, j})}, & a_{k, i} = j \in \{1, \dots, J_k\} \\[1mm]
        1 - p_{\text{d}}, & a_{k, i} = 0
    \end{cases} \nn \\[2mm]
    q(\underline{\V{x}}{}_{k, i}, 0, a_{k, i}; \V{z}_{k}) &\triangleq 1(a_{k, i}) \label{eq:def_q} \\[-3.3mm]
    \nn
\end{align}
where $1(a) \in \{0, 1\}$ is the indicator function of the event $a = 0$, i.e., $1(a) = 1$ if $a = 0$ and $0$ otherwise.

The factor $v(\overline{\V{y}}{}_{k, j}, b_{k, j}; \V{z}_{k, j}) \triangleq v(\overline{\V{x}}{}_{k, j}, \overline{r}{}_{k, j}, b_{k, j}; \V{z}_{k, j})$, describing the measurement model of the sensor as well as prior information for new \acp{po}, is given by
\begin{align}
    &v(\overline{\V{x}}{}_{k, j}, 1, b_{k, j}; \V{z}_{k, j}) \nn \\[1.5mm]
    & \hspace{4mm}\triangleq 
    \begin{cases}
        0, & b_{k, j} = i \in \{1, \dots, I_k\} \\[1.5mm]
        \frac{\displaystyle p_{\text{d}} \mu_{\text{u}} f_{\text{u}}(\overline{\V{x}}{}_{k, j}) f(\V{z}_{k, j} | \overline{\V{x}}{}_{k, j}) }{\displaystyle \mu_{\text{FA}} f_{\text{FA}} (\V{z}_{k, j})} , & b_{k, j} = 0
    \end{cases} \nn \\[3mm]
    &v(\overline{\V{x}}{}_{k, j}, 0, b_{k, j}; \V{z}_{k, j}) \triangleq f_{\text{D}}(\overline{\V{x}}{}_{k, j}) \label{eq:def_v}
\end{align}
where $f_{\text{D}}(\overline{\V{x}}{}_{k, j})$ is an arbitrary ``dummy'' \ac{pdf}. Here, the distribution $f_{\text{u}}(\overline{\V{x}}{}_{k, j})$ and mean number $\mu_{\text{u}}$ of unknown objects are used as prior information for new \acp{po}. Note that a detailed derivation of the factors $q(\underline{\V{y}}{}_{k, i}, a_{k, i}; \V{z}_{k})$ and $v(\overline{\V{y}}{}_{k, j}, b_{k, j}; \V{z}_{k, j})$ is provided in \cite{MeyKroWilLauHlaBraWin:J18}.

The factorization in \eqref{eq:factorization} provides the basis for a factor graph representation. Contrary to \cite{MeyKroWilLauHlaBraWin:J18}, in this work, we consider an alternative factor graph where PO states and association variables are combined in joint variable nodes. In particular, legacy PO states $\underline{\V{y}}{}_{k, i}$ and object-oriented association variables $a_{k, i}$ form joint nodes ``$\underline{\V{y}}{}_{k, i}, a_{k, i}$''. In addition, new PO states $\overline{\V{y}}{}_{k, j}$ and measurement-oriented association variables $b_{k, j}$ form joint nodes ``$\overline{\V{y}}{}_{k, j}, b_{k, j}$''. This combination of variable nodes is motivated by the fact that in the original factor graph there is exactly one $a_{k, i}$ connected to the corresponding $\underline{\V{y}}{}_{k, i}$ and exactly one $b_{k, j}$ connected to the corresponding $\overline{\V{y}}{}_{k, j}$. The resulting alternative factor graph leads to a presentation of the proposed method in Section \ref{sec:nebp_mot} that is consistent with \ac{bp} message passing rules \cite{KscFreLoe:01} as well as the original work on \ac{nebp} \cite{SatWel:21}. A single time step of the considered factor graph is shown in Fig.~\ref{fig:fg}. 

Next, \ac{bp} is applied to efficiently compute the beliefs $\tilde{f}(\V{y}_{k, n})$ that approximate the marginal posterior \ac{pdf} $f(\V{y}_{k, n} | \V{z}_{1 : k})$. Since the considered factor graph in Fig.~\ref{fig:fg} has loops, a specific message-passing order has to be chosen. As in \cite{MeyBraWilHla:J17, MeyKroWilLauHlaBraWin:J18}, we choose an order that is based on the following rules: (i) BP messages are only sent forward in time, and (ii) iterative message passing is only performed for data association and at each time step individually.

In what follows, we will briefly discuss the calculation of \ac{bp} messages on the considered factor graph shown in Fig.~\ref{fig:fg}. Note that messages sent from the singleton factor nodes ``$q(\underline{\V{y}}{}_{k, i}, a_{k, i}; \V{z}_{k})$'' to the joint variable nodes ``$\underline{\V{y}}{}_{k, i}, a_{k, i}$'', and messages sent from the singleton factor nodes ``$v(\overline{\V{y}}{}_{k, j}, b_{k, j}; \V{z}_{k, j})$'' to the joint variable nodes ``$\overline{\V{y}}{}_{k, j}, b_{k, j}$'' are equal to the singleton factor nodes ``$q(\underline{\V{y}}{}_{k, i}, a_{k, i}; \V{z}_{k})$'' and ``$v(\overline{\V{y}}{}_{k, j}, b_{k, j}; \V{z}_{k,j})$'' themselves. Thus, we reuse the same notation for factor nodes and corresponding messages.



\subsubsection{Prediction Messages}
The messages $\alpha_{k,i}(\underline{\V{y}}{}_{k, i}) \rmv\rmv= \alpha_{k,i}(\underline{\V{x}}{}_{k, i}, \underline{r}{}_{k, i})$, $i \rmv\in \{1, \dots, I_{k}\}$ passed from the factor nodes ``$f(\underline{\V{y}}{}_{k, i} | \V{y}{}_{k - 1, i})$'' to the joint variable nodes ``$\underline{\V{y}}{}_{k, i}, a_{k, i}$'' are calculated as (cf. \cite[Sec. IX-A1]{MeyKroWilLauHlaBraWin:J18})
\begin{equation}
    \alpha_{k,i}(\underline{\V{x}}{}_{k, i}, 1) = \int p_{\mathrm{s}} f(\underline{\V{x}}{}_{k, i} | \V{x}{}_{k - 1, i}) \tilde{f}(\V{x}{}_{k - 1, i}, 1) \mathrm{d} \V{x}{}_{k - 1, i}
\end{equation}
and $\alpha_{k,i}(\underline{\V{x}}{}_{k, i}, 0) = \alpha_{k, i} f_{\text{D}}(\underline{\V{x}}{}_{k, i})$ where $\alpha_{k, i}$ is a scalar that can be computed by making use of $\tilde{f}(\V{y}{}_{k - 1, i}) = \tilde{f}(\V{x}{}_{k - 1, i},r_{k - 1, i})$ and $p_{\mathrm{s}}$ (cf. \cite[eq. (77)]{MeyKroWilLauHlaBraWin:J18}). Note the belief $\tilde{f}(\V{y}{}_{k - 1, i})$ and the message $\alpha_{k, i}(\underline{\V{y}}{}_{k, i})$ are normalized in the sense that they sum to unity,\vspace{-2.5mm} e.g.,
\begin{equation}
    \sum_{\underline{r}{}_{k, i} \in \{0, 1\}} \int \alpha_{k,i}(\underline{\V{x}}{}_{k, i}, \underline{r}{}_{k, i}) \mathrm{d}\underline{\V{x}}{}_{k, i} = 1.
    \vspace{-1mm}
\end{equation}



\subsubsection{Iterative Probabilistic \ac{da}} \label{sec:iter_da}
At message passing iteration $\ell \in \{1, \dots, L\}$, the messages $\beta^{(\ell)}_{k, i,j}(\underline{\V{y}}{}_{k, i},a_{k, i})$ and $\xi^{(\ell)}_{k, j,i}(\overline{\V{y}}{}_{k, j},b_{k, j})$,  $i \rmv\in \{1, \dots, I_{k}\}$, $j \rmv\in\rmv \{1, \dots, J_{k}\}$ passed from variable nodes ``$\underline{\V{y}}{}_{k, i}, a_{k, i}$'' and ``$\overline{\V{y}}{}_{k, j}, b_{k, j}$''  to the indicator nodes ``$\Psi_{i, j} (a_{k, i}, b_{k, j})$'' are given by (cf.~\eqref{eq:bp_x_to_f})
\vspace{-1mm}
\begin{align}
\xi^{(\ell)}_{k, j,i}(\overline{\V{y}}{}_{k, j},b_{k, j}) &= v(\overline{\V{y}}{}_{k, j}, b_{k, j}; \V{z}_{k,j}) \prod_{\substack{i^\prime = 1 \\ i^\prime \ne i}}^{I_k} \rmv \varphi_{\Psi_{i^\prime, j} \to b_{k, j}}^{(\ell - 1)} \rmv\rmv (b_{k, j})\label{eq:def_xi2} \\[-2mm]
\beta^{(\ell)}_{k, i,j}(\underline{\V{y}}{}_{k, i},a_{k, i}) &= q(\underline{\V{y}}{}_{k, i}, a_{k, i}; \V{z}_{k}) \alpha_{k,i}(\underline{\V{y}}{}_{k, i}) \nn\\[1.5mm] &\hspace{15mm}\times \prod_{\substack{j^\prime = 1 \\ j^\prime \ne j}}^{J_k} \rmv\rmv \nu_{\Psi_{i, j^\prime} \to a_{k, i}}^{(\ell)} \rmv\rmv (a_{k, i}).
 \label{eq:def_beta2} \\[-5mm]
    \nn
\end{align}

In addition, the messages $\varphi_{\Psi_{i, j} \to b_{k, j}}^{(\ell)} (b_{k, j})$ passed from indicator nodes ``$\Psi_{i, j}(a_{k, i}, b_{k, j})$'' to variables nodes ``$\overline{\V{y}}{}_{k, i},b_{k, j}$'' and ``$\underline{\V{y}}{}_{k, i}, a_{k, i}$'' are obtained as (cf.~\eqref{eq:bp_f_to_x}),
\begin{align}
    \varphi_{\Psi_{i, j} \to b_{k, j}}^{(\ell)} (b_{k, j}) &= \sum_{a_{k, i} = 0}^{J_k} \Psi_{i, j}(a_{k, i}, b_{k, j}) \nn\\[-.5mm]
    &\hspace{6mm}\times \rmv\rmv\rmv\rmv\rmv \sum_{\underline{r}_{k, i} \in \{0,1\}} \int\beta^{(\ell)}_{k,i,j}(\underline{\V{x}}{}_{k, i},\underline{r}{}_{k, i},a_{k, i}) \mathrm{d} \underline{\V{x}}{}_{k, i} \nn\\[-3.5mm] 
    \label{eq:def_a_to_b1}\\
    \nu_{\Psi_{i, j} \to a_{k, i}}^{(\ell)} (a_{k, i}) &= \sum_{b_{k, j} = 0}^{I_k} \Psi_{i, j}(a_{k, i}, b_{k, j}) \nn\\
    &\hspace{4mm}\times \rmv\rmv\rmv\rmv\rmv  \sum_{\overline{r}_{k, \rd{j}} \in \{0,1\}} \int \xi^{(\ell)}_{k, j,i}(\overline{\V{x}}{}_{k, j},\overline{r}{}_{k, j},b_{k, j}) \mathrm{d} \overline{\V{x}}{}_{k, j}. \nn\\[-3.5mm] 
    \label{eq:def_b_to_a1}
\end{align}

By plugging \eqref{eq:def_xi2} into \eqref{eq:def_b_to_a1} and \eqref{eq:def_beta2} into \eqref{eq:def_a_to_b1}, we finally obtain the following combined\vspace{1mm} messages for  $i \rmv\in \{1, \dots, I_{k}\}$, and $j \rmv\in\rmv \{1, \dots, J_{k}\}$
\begin{align}
    &\varphi_{\Psi_{i, j} \to b_{k, j}}^{(\ell)} (b_{k, j}) \nn \\[2.2mm]
    &\hspace{3mm}= \sum_{a_{k, i} = 0}^{J_k} \beta_{k,i}(a_{k, i}) \Psi_{i, j}(a_{k, i}, b_{k, j}) \prod_{\substack{j^\prime = 1 \\ j^\prime \ne j}}^{J_k} \nu_{\Psi_{i, j^\prime} \to a_{k, i}}^{(\ell)} (a_{k, i}) \label{eq:def_a_to_b} \\
    &\nu_{\Psi_{i, j} \to a_{k, i}}^{(\ell)} (a_{k, i}) \nn \\[2.2mm]
    &\hspace{3mm}= \sum_{b_{k, j} = 0}^{I_k} \xi_{k,j}(b_{k, j}) \Psi_{i, j}(a_{k, i}, b_{k, j}) \prod_{\substack{i^\prime = 1 \\ i^\prime \ne i}}^{I_k} \varphi_{\Psi_{i^\prime, j} \to b_{k, j}}^{(\ell - 1)} (b_{k, j}). \label{eq:def_b_to_a}
\end{align}
Here, we have introduced the short notation
\begin{align}
&\beta_{k,i}(a_{k, i}) \triangleq \sum_{\underline{r}_{k, i} \in \{0,1\}} \ist\ist \int \ist q(\underline{\V{x}}{}_{k, i}, \underline{r}{}_{k, i}, a_{k, i}; \V{z}_{k}) \nn\\
&\hspace{40.8mm}\times \alpha_{k,i}(\underline{\V{x}}{}_{k, i}, \underline{r}{}_{k, i}) \mathrm{d} \underline{\V{x}}{}_{k, i} \label{eq:beta}
 \end{align}
and 
\vspace{.5mm}
\begin{equation}
\xi_{k,j}(b_{k, j})\triangleq\rmv \sum_{\overline{r}_{k, j} \in \{0,1\}} \int v(\overline{\V{x}}{}_{k, j}, \overline{r}{}_{k, j}, b_{k, j}; \V{z}_{k,j}) \mathrm{d} \overline{\V{x}}{}_{k, j}. \label{eq:xi}
\end{equation}
Note that with a small abuse of notation, one message passing iteration, indexed by $\ell$, in \eqref{eq:def_a_to_b} and \eqref{eq:def_b_to_a} corresponds to two message passing iterations in \eqref{eq:bp_x_to_f} and \eqref{eq:bp_f_to_x}. Iterative message passing \eqref{eq:def_a_to_b} and \eqref{eq:def_b_to_a}  is initialized at $\ell \rmv=\rmv 1$, by setting in \eqref{eq:def_b_to_a} $\varphi_{\Psi_{i, j} \to b_{k, j}}^{(0)} (b_{k, j}) \rmv=\rmv 1$ for all $j \in \{1,\dots,J_k\}$ and $i \in \{1,\dots,$ $I_k\}$.

These combined messages can be further simplified \cite{WilLau:J14} as follows. Because of the binary consistency constraints expressed by $\Psi_{i, j}(a_{k, i}, b_{k, j})$, each message comprises only two different values. In particular, $\varphi_{\Psi_{i, j} \to b_{k, j}}^{(\ell)} (b_{k, j})$ in \eqref{eq:def_a_to_b} takes on one value for $b_{k, j} = i$ and another for all $b_{k, j} \ne i$. Furthermore, $\nu_{\Psi_{i, j} \to a_{k, i}}^{(\ell)} (a_{k, i})$ in \eqref{eq:def_b_to_a} takes on one value for $a_{k, i} = j$ and another for all $a_{k, i} \ne j$. Thus, each message can be represented (up to an irrelevant constant factor) by the ratio of the first value and the second value, hereafter denoted as $\varphi_{i, j}^{(\ell)}$ for $\varphi_{\Psi_{i, j} \to b_{k, j}}^{(\ell)} (b_{k, j})$ and $\nu_{j, i}^{(\ell)}$ for $\nu_{\Psi_{i, j} \to a_{k, i}}^{(\ell)} (a_{k, i})$. By exchanging simplified messages the computational complexity of each message passing iteration only scales as $\mathcal{O}(I_k J_k)$ (see \cite{WilLau:J14,MeyKroWilLauHlaBraWin:J18} for details). Furthermore, it has been shown in \cite{WilLau:J14} that iterative probabilistic data association following \eqref{eq:def_a_to_b}--\eqref{eq:def_b_to_a} and its simplified version discussed above are guaranteed to converge.

\vspace{2mm}



\subsubsection{Belief Calculation} \label{sec:belief_cal}
Finally, after the last message passing iteration $\ell = L$, the beliefs $\tilde{f}(\underline{\V{y}}{}_{k, i}, a_{k, i})$, $i \rmv\in\rmv\{1,\dots,I_{k}\}$ and $\tilde{f}(\overline{\V{y}}{}_{k, j}, b_{k, j})$, $j \rmv\in\rmv\{1,\dots, J_{k}\}$ are computed according\vspace{-2mm} to
\begin{align}
    &\tilde{f}(\underline{\V{y}}{}_{k, i}, a_{k, i}) = \frac{1}{\underline{C}{}_{k, i}} \ist q(\underline{\V{y}}{}_{k, i}, a_{k, i}; \V{z}_{k}) \ist \alpha_{k,i}(\underline{\V{y}}{}_{k, i}) \nn \\
    & \hspace{51.2mm} \times \prod_{j = 1}^{J_k} \nu_{\Psi_{i, j} \to a_{k, i}}^{(L)} (a_{k, i}) \nn\\[-1mm]
    &\tilde{f}(\overline{\V{y}}{}_{k, j}, b_{k, j}) = \frac{1}{\overline{C}{}_{k, j}} \ist v(\overline{\V{y}}{}_{k, j}, b_{k, j}; \V{z}_{k, j}) \prod_{i = 1}^{I_k} \varphi_{\Psi_{i, j} \to b_{k, j}}^{(L)} (b_{k, j}) \nn \\[-5mm]
    \nn
\end{align}
where $\underline{C}{}_{k, i}$, and $\overline{C}{}_{k, j}$ are normalizing constants that make sure that $\tilde{f}(\underline{\V{y}}{}_{k, i}, a_{k, i})$ and $\tilde{f}(\overline{\V{y}}{}_{k, j}, b_{k, j})$ sum and integrate to unity. The marginal beliefs $\tilde{f}(\underline{\V{y}}{}_{k, i})$, $\tilde{f}(\overline{\V{y}}{}_{k, j})$, $\tilde{p}(a_{k, i})$, and $\tilde{p}(b_{k, j})$ can then be obtained from $\tilde{f}(\underline{\V{y}}{}_{k, i}, a_{k, i})$ and $\tilde{f}(\overline{\V{y}}{}_{k, j}, b_{k, j})$ by marginalization. In particular, the approximate marginal posterior \acp{pdf} of augmented states $\tilde{f}(\underline{\V{y}}{}_{k, i}) \rmv=\rmv f(\underline{\V{y}}{}_{k, i}| \V{z}_{1: k})$, $i \rmv\in\rmv\{1,\dots, I_{k}\}$ and $\tilde{f}(\overline{\V{y}}{}_{k, j}) \rmv=\rmv f(\overline{\V{y}}{}_{k, j}| \V{z}_{1: k})$, $j \rmv\in\rmv\{1,\dots,J_{k}\}$ are used for object declaration and state estimation as discussed in Section \ref{sec:estimationDetection}. Furthermore,  the approximate marginal association probabilities $\tilde{p}(a_{k, i}) \rmv=\rmv p(a_{k, i}| \V{z}_{1: k})$, $i \rmv\in\rmv\{1,\dots, I_{k}\}$ and $\tilde{p}(b_{k, j}) \rmv=\rmv p(b_{k, j}| \V{z}_{1: k})$, $j \rmv\in\rmv\{1,\dots,J_{k}\}$ are used in a preprocessing step for performance evaluation discussed in Sections \ref{sec:experimentalSetup} and \ref{subsec:exp_imp}.

\vspace{-1mm}



\section{Proposed NEBP-based MOT Algorithm} \label{sec:nebp_mot}
In this section, we start with a discussion on how neural networks can extract features from raw sensor data by using previous state estimates and preprocessed measurements. We then introduce the proposed \ac{nebp} framework for \ac{mot}, which compared to \ac{bp} for MOT uses features as an additional input. Since we limit our discussion to a single time step, we will omit the time index $k$ in what\vspace{-4mm} follows.



\subsection{Feature Extraction} \label{sec:feat_extract}

We consider two types of features: (i) features that represent motion information (e.g., position and velocity) and (ii) features that represent shape information. 

For each legacy \ac{po} $i \rmv=\rmv \big\{1,\dots,I\big\}$, the shape feature $\V{h}_{a_i, \text{shape}}$ is extracted\vspace{-1.5mm} as
\begin{equation}
    \V{h}_{a_i, \text{shape}} = g_{\text{shape}, 1}(\Set{Z}^{-}\rmv; \underline{\hat{\V{x}}}{}_{i}^{-})
    \vspace{.2mm} \label{eq:feat_shape_a}
 \end{equation}
 where $\hat{\underline{\V{x}}}_i^{-}$ is approximate \ac{mmse} state estimate of legacy \ac{po} $i$ at the previous time step. Furthermore, $\Set{Z}^{-}$ is the raw sensor data at the previous time step and $g_{\text{shape}, 1}(\Set{Z}^{-}\rmv; \underline{\hat{\V{x}}}{}_{i}^{-})$ is a neural network.

Similarly, for each preprocessed measurement $\V{z}_{j}$, $j \rmv=\rmv \big\{1,\dots,J\big\}$ the shape feature $\V{h}_{b_j, \text{shape}}$ is obtained\vspace{-1mm} as
\begin{equation}
\V{h}_{b_j, \text{shape}} = g_{\text{shape}, 2}(\Set{Z}; \V{z}_{j}) \label{eq:feat_shape_b}
\end{equation}
where $g_{\text{shape}, 2}(\Set{Z}; \V{z}_{j})$ is again a neural network and $\Set{Z}$ is the raw sensor data collected at the current time step.

Finally, for each legacy \ac{po} $i \rmv=\rmv \big\{1,\dots,I\big\}$ and each measurement $j \rmv=\rmv \big\{1,\dots,J\big\}$, a motion feature is computed according\vspace{-2.5mm} to
\begin{align}
    \V{h}_{a_i, \text{motion}} &= g_{\text{motion}, 1}(\underline{\hat{\V{x}}}{}_{i}^{-},\underline{\hat{r}}{}_{i}^{-}) \nn\\[1mm]
    \V{h}_{b_j, \text{motion}} &= g_{\text{motion}, 2}(\V{z}_{j}) \label{eq:feat_motion}
\end{align}
where $\underline{\hat{r}}{}_{i}^{-}$ is the approximate existence probability of legacy \ac{po} $i$. Furthermore, $g_{\text{motion}, 1}(\underline{\hat{\V{x}}}{}_{i}^{-},\underline{\hat{r}}{}_{i}^{-})$ and $g_{\text{motion}, 2}(\V{z}_{j})$ are again neural networks. We will discuss one particular instance of shape feature extraction in\vspace{-2mm} Section \ref{subsec:exp_imp}.



\subsection{GNN Topology and BP Message Enhancement} \label{subsec:gnn_topo_nebp}

The conjecture of this work is that in many \ac{mot} applications (i) object dynamics and existence can be described accurately by a statistical model represented by the \acp{pdf} $f(\underline{\V{x}}{}_{k, i} | \V{x}_{k - 1, i})$, $f_{\text{u}}(\overline{\V{x}}{}_{k, j})$ and parameters $p_{\text{s}}$, $\mu_{\text{u}}$; (ii) measurement detection and the resulting measurements of the object's position can also be described well by a statistical model represented by \acp{pdf} $f(\V{z}_{k, j} | \V{x}_{k, n})$, $f_{\text{FA}} (\V{z}_{k, j})$ and parameters $p_{\text{d}}$, $\mu_{\text{FA}}$; but (iii) object shape information are difficult to represent accurately by a statistical model. Thus, we can make use of models available for (i) and (ii), but for (iii), we best learn the influence of object shape information on measurement detection directly from the data itself. Thus, contrary to the original \ac{nebp} approach, in our \ac{nebp} method, only the parts of the \ac{mot} factor graph that model the data generating process are matched by the \ac{gnn}. These matched parts are highlighted in Fig.~\ref{fig:fg_gnn}. All factor nodes in this part of the factor graph are either singleton or pairwise factor nodes. As discussed in Section \ref{subsec:nebp}, in \ac{nebp} singleton factor nodes are not matched by \ac{gnn} nodes. In addition, in our considered factor graph in Fig.~\ref{fig:fg_gnn}(a), the pairwise factor nodes ``$\Psi_{i, j}(a_{i}, b_{j})$''\rmv, $j \rmv=\rmv \big\{1,\dots,J\big\}$ and $i \rmv=\rmv \big\{1,\dots,I\big\}$ represent simple binary consistency constraints. Thus, we do not explicitly model factor nodes by \ac{gnn} nodes. The node embeddings of \acp{gnn} nodes introduced for the variable nodes ``$\underline{\V{y}}{}_i, a_{i}$'', $i \rmv=\rmv \big\{1,\dots,I\big\}$ are denoted as $\V{h}_{a_{i}}$ and the node embeddings introduced for variables nodes ``$\overline{\V{y}}{}_j, b_{j}$'', $j \rmv=\rmv \big\{1,\dots,J\big\}$ are denoted as $\V{h}_{b_j}$. Finally, following the topology of the factor graph for data association in Fig.~\ref{fig:fg_gnn}(a), GNN edges are introduced such that the bipartite GNN shown in Fig.~\ref{fig:fg_gnn}(b) is obtained.  

We recall from Section~\ref{subsec:fg_bp} that for singleton factor nodes, the message passed to the adjacent variable node is equal to the factor node itself. As a result, $q(\underline{\V{x}}{}_{i}, \underline{r}_i, a_{i}; \V{z})$  and $v(\overline{\V{x}}{}_{j}, \overline{r}_j, b_{j}; \V{z}_j)$ not only describe factor nodes but also the messages that are enhanced. There are two challenges related to directly enhancing these messages based on the GNN according to \eqref{eq:nebp_x_to_f}--\eqref{eq:nebp_ori_comb}, i.e., (i) the codomain of $q(\underline{\V{x}}{}_{i}, \underline{r}_{i}, a_{i}; \V{z})$ and $v(\overline{\V{x}}{}_{j}, \overline{r}_{j}, b_{j}; \V{z}_j)$ can be very large, which complicates the training of the GNN \cite{IofSze:15} (see also Sections \ref{sec:statementMOT} and  \ref{sec:loss}) and (ii) the messages $q(\underline{\V{x}}{}_{i}, \underline{r}_{i}, a_{i}; \V{z})$ and $v(\overline{\V{x}}{}_{j}, \overline{r}{}_{j}, b_{j}; \V{z}_j)$ involve the continuous random variables $\underline{\V{x}}{}_{i}$ and $\overline{\V{x}}{}_{j}$ which makes it impossible to enhance them by the output of a \ac{gnn} for every possible value of $\underline{\V{x}}{}_{i}$ and $\overline{\V{x}}{}_{j}$ individually.

To address the first challenge, we introduce normalized versions\footnote{Multiplying BP messages by a constant factor does not alter the resulting beliefs \cite{KscFreLoe:01}.} of the original BP messages\vspace{-.8mm} as
\begin{align}
    q_\mathrm{s}(\underline{\V{x}}{}_i, \underline{r}{}_i, a_{i}; \V{z}) &= \frac{1}{C_q} q(\underline{\V{x}}{}_i, \underline{r}{}_i, a_{i}; \V{z}) \label{eq:qs}\\[1mm]
    v_\mathrm{s}(\overline{\V{x}}{}_j, \overline{r}{}_i, b_{j}; \V{z}_j) &= \frac{1}{C_v} v(\overline{\V{x}}{}_j, \overline{r}{}_i, b_{j}; \V{z}_j) \nn \\[-7.5mm]
    \label{eq:vs}
\end{align}
where $C_q = \sum_{a_{i} = 0}^J \sum_{\underline{r}{}_i \in \{0, 1\}} \int q(\underline{\V{x}}{}_i, \underline{r}{}_i, a_{i}; \V{z}) \mathrm{d}\underline{\V{x}}{}_i$ and $C_v = \sum_{b_{j} = 0}^I \sum_{\overline{r}{}_i \in \{0, 1\}} \int v(\overline{\V{x}}{}_j, \overline{r}{}_i, b_{j}; \V{z}_j) \mathrm{d}\overline{\V{x}}{}_i$ \vspace{.3mm} are the normalization constants. Note that after normalization the codomain of $q_\mathrm{s}(\underline{\V{x}}{}_{i}, \underline{r}_{i}, a_{i}; \V{z})$ and $v_\mathrm{s}(\overline{\V{x}}{}_{j}, \overline{r}_{j}, b_{j}; \V{z}_j)$ is limited to the interval $[0,1]$.

The second challenge is addressed by enhancing \ac{bp} messages $q_\mathrm{s}(\underline{\V{x}}{}_i, \underline{r}{}_i, a_{i}; \V{z})$, $i \in \{1, \dots, I\}$ and $v_\mathrm{s}(\overline{\V{x}}{}_j, \overline{r}{}_i, b_{j}; \V{z}_j)$, $j \in \{1, \dots, J\}$ as follows (cf.~\eqref{eq:nebp_ori_comb})
\begin{align}
    \tilde{q}_{\mathrm{s}}(\underline{\V{x}}{}_i, 1, a_{i} = j; \V{z}) &= \omega_j \cdot q_{\mathrm{s}}(\underline{\V{x}}{}_i, 1, a_{i} = j; \V{z}) + \mu_{i}(j) \label{eq:nebp_combine_a} \\[1mm]
    \tilde{v}_{\mathrm{s}}(\overline{\V{x}}{}_j, 1, b_{j} = 0; \V{z}_j) &= \omega_j \cdot v_{\mathrm{s}}(\overline{\V{x}}{}_j, 1, b_{j} = 0; \V{z}_j). \label{eq:nebp_combine_b}
\end{align}
Here, $\omega_j \in (0, 1)$ and $\mu_{i}(j) \in \mathbb{R}^+$ are computed from information provided by the GNN as discussed in the following Section~\ref{sec:statementMOT}.  The other entries of the messages $\tilde{q}_{\mathrm{s}}(\underline{\V{x}}{}_i, \underline{r}_i, a_{i}; \V{z})$, $i \in \{1, \dots, I\}$ and $\tilde{v}_{\mathrm{s}}(\overline{\V{x}}{}_j, \overline{r}_j, b_{j}; \V{z}_j)$, $j \in \{1, \dots, J\}$ remain
unenhanced, i.e., $\tilde{q}_{\mathrm{s}}(\underline{\V{x}}{}_i, 0, a_{i}; \V{z}) \rmv=\rmv q_{\mathrm{s}}(\underline{\V{x}}{}_i, 0, a_{i}; \V{z})$, $\tilde{q}_{\mathrm{s}}(\underline{\V{x}}{}_i, 1, a_{i} \rmv=\rmv 0; \V{z}) = q_{\mathrm{s}}(\underline{\V{x}}{}_i, 1, a_{i} \rmv=\rmv 0; \V{z})$, $\tilde{v}_{\mathrm{s}}(\overline{\V{x}}{}_j, 0, b_{j}; \V{z}_j) = v_{\mathrm{s}}(\overline{\V{x}}{}_j, 0, b_{j}; \V{z}_j)$, and $\tilde{v}_{\mathrm{s}}(\overline{\V{x}}{}_j, 1, b_{j} = i; \V{z}_j) = v_{\mathrm{s}}(\overline{\V{x}}{}_j, 1, b_{j} = i; \V{z}_j), i \in \{1, \dots, I\}$.  Note that calculating \ac{nebp} messages according to \eqref{eq:nebp_combine_a} and \eqref{eq:nebp_combine_b} avoids enhancing $q(\underline{\V{x}}{}_{i}, \underline{r}_i, a_{i}; \V{z})$ and $v(\overline{\V{x}}{}_{j}, \overline{r}_j, b_{j}; \V{z}_j)$ for every possible value of $\underline{\V{x}}{}_i$ and $\overline{\V{x}}{}_j$. All other BP messages are not enhanced. 

Finally, neural enhanced data association can be performed by replacing the functions $\beta_i(a_{i})$ and $\xi_j(b_{j})$ in \eqref{eq:def_a_to_b} and \eqref{eq:def_b_to_a} with their neural enhanced counterparts $\tilde{\beta}_{\mathrm{s},i}(a_{i})$ and $\tilde{\xi}_{\mathrm{s},j}(b_{j})$. These neural enhanced counterparts are obtained by replacing in \eqref{eq:beta} and \eqref{eq:xi} the \ac{bp} messages $q(\underline{\V{x}}{}_{i}, \underline{r}{}_{i}, a_{i}; \V{z})$ and $v(\overline{\V{x}}{}_{j}, \overline{r}{}_{j}, b_{j}; \V{z}_{j})$ with the \ac{nebp} messages $\tilde{q}_\mathrm{s}(\underline{\V{x}}{}_{i}, \underline{r}{}_{i}, a_{i}; \V{z})$ and $\tilde{v}_\mathrm{s}(\overline{\V{x}}{}_{j}, \overline{r}{}_{j}, b_{j}; \V{z}_{j})$, respectively. In particular, for $i \in \{1, \dots, I\}$ we\vspace{-.5mm} obtain 
\begin{equation}
    \tilde{\beta}_{\mathrm{s},i}(a_{i} = j) = \omega_j \frac{\beta_i(a_{i} = j)}{C_q} + \mu_{i}(j) \quad j \in \{1, \dots, J\} \label{eq:beta_enhanced}
    \vspace{-.5mm}
\end{equation}
and $\tilde{\beta}_{\mathrm{s},i}(a_{i} = 0) = \beta_i(a_{i} = 0)$. Similarly, for $j \in \{1, \dots, J\}$ we\vspace{-.5mm} get 
\begin{equation}
    \tilde{\xi}_{\mathrm{s},j}(b_{j} = 0) = \omega_j \frac{\xi_j(b_{j} = 0)}{C_v}  \label{eq:xi_enhanced}
    \vspace{-.5mm}
\end{equation}
and $\tilde{\xi}_{\mathrm{s},j}(b_{j} = i) = \xi_{\mathrm{s},j}(b_{j} = i)$, $i \in \{1, \dots, I\}$.


The value $\beta_i(a_{i} = j)$, $j \rmv\in\rmv \{1,\dots,J\}$ provides a likelihood ratio for the measurement with index $j$ being associated to the legacy \ac{po} with index $i$ \cite{MeyKroWilLauHlaBraWin:J18}. In addition, $\xi_j(b_{j} = 0)$ provides a likelihood ratio for the measurement with index $j$ being generated by a new \ac{po}. The shape association term $\mu_{i}(j) \ge 0$ in \eqref{eq:beta_enhanced}, calculated by the \ac{gnn} implements object shape association, which can be interpreted as follows. The GNN compares the shape feature extracted for legacy PO $j$ with the shape feature extracted for measurement $j$ and, if there is a good match, outputs a large $\mu_{i}(j) > 0$. According to \eqref{eq:beta_enhanced}, this effectively increases the likelihood ratio that the legacy \ac{po} $i$ is associated with the measurement $j$. Note that there is no shape association term in \eqref{eq:xi_enhanced}. Since the shape feature extracted for new \ac{po} $j$ would be the same as the shape feature for measurement $j$, comparing shape features as performed for legacy POs and measurements is not possible.

The scalar $\omega_{j} \rmv\rmv\in\rmv\rmv (0, 1) $ in \eqref{eq:beta_enhanced} and \eqref{eq:xi_enhanced} provided by the \ac{gnn} implements false alarm rejection. In particular, $\omega_{j} < 1$ is equal to the local increase of the false alarm distribution according to $\tilde{f}_{\text{FA}}(\V{z}_{j}) = \frac{1}{\omega_{j}} f_{\text{FA}}(\V{z}_{j})$ (cf.~\eqref{eq:qs}, \eqref{eq:vs}, \eqref{eq:def_q}, and \eqref{eq:def_v}). In \eqref{eq:beta_enhanced}, this local increase of the false alarm distribution makes it less likely that the measurement $\V{z}_{j}$ is associated to a legacy PO. In \eqref{eq:xi_enhanced}, this local increase reduces the existence probability of the new PO introduced for the\vspace{-1mm} measurement $\V{z}_{j}$.

\subsection{Statement of the \ac{nebp} for \ac{mot} Algorithm}
\label{sec:statementMOT}
 \ac{nebp} for \ac{mot} consists of the following steps:

    \subsubsection{Conventional \ac{bp}}
    First, the conventional \ac{bp}-based \ac{mot} algorithm is run until convergence, from which we obtain $\beta_{\mathrm{s}, i} = [\beta_{\mathrm{s}, i}(0) \cdots \beta_{\mathrm{s}, i}(J)]^\T \in \mathbb{R}^{J + 1}, \xi_{\mathrm{s}, j} = [\xi_{\mathrm{s}, j}(0) \cdots \xi_{\mathrm{s}, j}(I)]^\T \in \mathbb{R}^{I + 1}, \varphi_{i, j} \in \mathbb{R}$, and $\nu_{i, j} \in \mathbb{R}$ (cf. Section \ref{sec:iter_da}), where $\beta_{\mathrm{s}, i}(j) = \beta_i(a_{i} = j) / C_q$ and $\xi_{\mathrm{s}, j}(i) = \xi(b_{j} = i) / C_{v}$.


    \subsubsection{\ac{gnn} Messages}
    Next, \ac{gnn} message passing is\vspace{.5mm} executed iteratively. In particular, at iteration $p \rmv\in\rmv \{1,\dots,P\}$ the messages passed along the edges of \ac{gnn} are computed\vspace{.5mm} as
    \begin{align}
    	\V{m}_{a_i \to b_j}^{(p)} &= g_{\text{e}} \Big( \V{h}_{a_i}^{(p)}, \V{h}_{b_j}^{(p)}, \beta_{\mathrm{s}, i}(j), \varphi_{i, j} \Big) \label{eq:gnn_a_to_b} \\[1mm]
    	\V{m}_{b_j \to a_i}^{(p)} &= g_{\text{e}} \Big(\V{h}_{a_i}^{(p)}, \V{h}_{b_j}^{(p)}, \beta_{\mathrm{s}, i}(j), \nu_{i, j} \Big)\label{eq:gnn_a_to_b2}
	\end{align}
where $g_{\text{e}}(\cdot)$ is the edge neural network. Furthermore, node embedding vectors of each node are obtained as 
	\begin{align}
    	\V{h}_{a_i}^{(p+1)} &= g_{\text{n}} \bigg(\V{h}_{a_i}^{(p)}, \sum_{j \in \mathcal{N}(i)} \V{m}_{b_j \to a_i}^{(p)}, \beta_{\mathrm{s}, i}(0) \bigg) \label{eq:gnn_a_to_b3} \\	
    	\V{h}_{b_j}^{(p+1)} &= g_{\text{n}} \bigg(\V{h}_{b_j}^{(p)}, \sum_{j \in \mathcal{N}(i)} \V{m}_{a_i \to b_j}^{(p)}, \xi_{\mathrm{s}, j}(0) \bigg). \label{eq:gnn_b}
    \end{align}
     where $g_{\text{n}}(\cdot)$ is the node neural network. The iterative processing scheme is initialized by setting node embeddings equal to motion and shape features, i.e., $\V{h}_{a_i}^{(1)} \rmv=\rmv [\V{h}_{a_i, \text{motion}}^\T$ $\V{h}_{a_i, \text{shape}}^\T]^\T$ and $\V{h}_{b_j}^{(1)} \rmv=\rmv [\V{h}_{b_j, \text{motion}}^\T$ $\V{h}_{b_j, \text{shape}}^\T]^\T\rmv$. 
    \vspace{1.5mm}
    
    \subsubsection{NEBP Messages}
    After computing \eqref{eq:gnn_a_to_b}-\eqref{eq:gnn_b} for $P$ iterations, the refinement $\omega_j$ used in \eqref{eq:nebp_combine_a} and \eqref{eq:nebp_combine_b} is calculated\vspace{-1mm} as
    	\begin{equation}
        \omega_j = \sigma \big( T(\omega^{*}_j - \delta) \big) \text{ with } \omega_j^{*} = g_{\mathrm{s}}\big(\V{h}_{b_j}^{(P)}\big).
        \label{eq:nebp_omega}
    \end{equation}
Here, $g_{\mathrm{s}}(\cdot)$ is a neural network and $\sigma(x) = 1 / (1 + e^{-x}) \in (0,1)$ is the sigmoid function. Furthermore, the temperature $T$ and the bias $\delta$ are hyperparameters \cite{GuoPleSunWei:17} that make it possible to calibrate the transition of the sigmoid.       
         
         Finally, the refinement $\mu_{i}(j)$ used in \eqref{eq:nebp_combine_a} is obtained as 
\begin{equation}
        \mu_{i}(j) = \text{ReLU}\big( \mu^{*}_{i}(j) \big) \text{ with } \mu^{*}_{i}(j) = g_{\text{d}}\big(\V{m}_{b_j \to a_i}^{(P)} \big)
        \label{eq:nebp_mu}
    \end{equation}
    where $g_{\text{d}}(\cdot)$ is another neural network and $\text{ReLU}(\cdot)$ is the rectified linear unit.

    \subsubsection{Belief Calculation}
    Finally, iterative probabilistic \ac{da} is again run until convergence by replacing $q_{\mathrm{s}}(\underline{\V{x}}{}_{i}, \underline{r}{}_{i}, a_{i}; \V{z})$ and $v_{\mathrm{s}}(\overline{\V{x}}{}_{j}, \overline{r}{}_{i}, b_{j}; \V{z}_j)$ in \eqref{eq:qs} and \eqref{eq:vs} by its neural enhanced counterparts $\tilde{q}_{\mathrm{s}}(\underline{\V{x}}{}_{i}, \underline{r}{}_{i}, a_{i}; \V{z})$ and $\tilde{v}_{\mathrm{s}}(\overline{\V{x}}{}_{j}, \overline{r}{}_{i}, b_{j}; \V{z}_j)$ in \eqref{eq:nebp_combine_a} and \eqref{eq:nebp_combine_b}, respectively. This results in the enhanced\vspace{-0mm} messages $\tilde{\varphi}_{\Psi_{i, j} \to b_{j}}^{(L)} (b_{j})$ and $\tilde{\nu}_{\Psi_{i, j} \to a_{i}}^{(L)} (a_{i})$ (cf. Section \ref{sec:iter_da}), which are then used for the calculation of legacy \ac{po} beliefs $\tilde{f}(\underline{\V{y}}{}_{i})$, $i \in \{1, \dots, I\}$ and new \ac{po} beliefs $\tilde{f}(\overline{\V{y}}{}_{j}), j \in \{1, \dots, J\}$ as discussed in Section \ref{sec:belief_cal}.
\vspace{-2.5mm}

\subsection{Complexity Analysis} \label{subsec:complexity-analysis}
In this section, we analyze and compare the computational complexity of the conventional \ac{bp} and the proposed \ac{nebp} methods for \ac{mot}. Since both \ac{bp} and \ac{nebp} for \ac{mot} follow the detect-then-track paradigm, the complexity of the detector has to be taken into account. We denote by $|\Set{Z}|$ the number of raw sensor data points. For example, if a LiDAR sensor is considered, this is the number of points of the LiDAR point cloud, and if a camera sensor is used, this is the number of pixels of the camera image. Then the number of operations needed for detection is $c_{\text{det}}|\Set{Z}|$, where $c_{\text{det}}$ is a constant that depends on the size and type of neural network used as the detector $g_{\text{det}}(\cdot)$. As discussed in \cite{MeyBraWilHla:J17, MeyKroWilLauHlaBraWin:J18}, the number of operations needed for the conventional \ac{bp} method for \ac{mot} algorithm is $c_{\text{bp}} \ist I \ist J$, where $c_{\text{bp}}$ is a constant that depends on the number of message passing iterations for \ac{da}, the number of particles, and further parameters. In total, the number of operations for \ac{bp} is $c_{\text{det}}|\Set{Z}| + c_{\text{bp}} \ist IJ$. Thus, the computational complexity of \ac{bp} scales as $\Set{O}(|\Set{Z}| + IJ)$.

The additional operations of \ac{nebp} compared to conventional \ac{bp} are related to feature extraction and the \ac{gnn}. Feature extraction as discussed in \eqref{eq:feat_shape_a}--\eqref{eq:feat_motion} requires $c_{\text{shape}}|\Set{Z}| + c_{\text{motion}}(I + J)$ operations, where $c_{\text{shape}}, c_{\text{motion}}$ are constants that depend on the size and type of the neural networks $g_{\text{shape}}(\cdot), g_{\text{motion}}(\cdot)$, respectively. The \ac{gnn} is a fully connected bipartite graph, i.e., it consists of two sets of nodes, and each node in the first set is connected via an edge to each node in the second set. The number of nodes in each set is equal to $I$ and $J$, respectively. GNN messages are exchanged on the $IJ$ edges of the network according to \eqref{eq:gnn_a_to_b} and \eqref{eq:gnn_a_to_b2}. This is followed by GNN messages aggregation in \eqref{eq:gnn_a_to_b3} and \eqref{eq:gnn_b}, as well as BP message refinement in \eqref{eq:nebp_omega} and \eqref{eq:nebp_mu}. The total number of operations is hence equal to $c_{\text{gnn}, 1}IJ + c_{\text{gnn}, 2}I + c_{\text{gnn}, 3}J$, where $c_{\text{gnn}, \cdot}$ depends on the size and type of neural networks $g_{\text{e}}(\cdot)$, $g_{\text{n}}(\cdot)$, $g_{\text{s}}(\cdot)$,  and $g_{\text{d}}(\cdot)$. It can thus be seen, that the computational complexity of \ac{nebp} also scales as $\Set{O}(|\Set{Z}| + IJ)$. Note that due to the additional operations performed by \ac{nebp}, the runtime of \ac{nebp} is longer compared to \ac{bp}. Runtimes of \ac{bp} and \ac{nebp} are further analyzed in Section~\ref{subsec:performance}.



\section{Loss Function and Training} \label{sec:loss}

Training of the proposed \ac{nebp} method is performed in a supervised manner. It is assumed that a training set consisting of ground truth object tracks is available. A ground truth object track is a sequence of object positions. Every sequence is characterized by an object identity (ID). During the training phase, the parameters of all neural networks are updated through back-propagation, which computes the gradient of the loss function. The loss function has the form $\Set{L} = \Set{L}_{\mathrm{r}} + \Set{L}_{\mathrm{a}}$, where the two contributions $\Set{L}_{\mathrm{r}}$ and $\Set{L}_{\mathrm{a}}$ are related to false alarm rejection and object shape association, respectively. 

Thus, we consider the following binary cross-entropy loss \cite[Chapter 4.3]{Bishop:B06} for false alarm rejection, i.e.,\vspace{0mm}
\begin{equation}
    \Set{L}_{\text{r}} = \frac{-1}{J} \sum_{j = 1}^{J}  \omega_j^{\text{gt}} \ln(\omega_{j}) + \epsilon (1 - \omega_j^{\text{gt}}) \ln(1 - \omega_{j}) \label{eq:loss_omega} \vspace{0mm}
\end{equation}
where $\omega_j^{\text{gt}} \rmv\in\rmv \{0,1\}$ is the pseudo ground truth label for each measurement and $\epsilon \rmv\in\rmv \mathbb{R}^{+}$ is a tuning parameter. The pseudo ground truth label $\omega_j^{\text{gt}}$ is equal to $1$ if the distance between the measurement and any ground truth position is smaller or equal to $T_{\text{dist}}$, and $0$ otherwise.  The tuning parameter $\epsilon \rmv\in\rmv \mathbb{R}^{+}$ addresses the imbalance problem in learning-based binary classification (see \cite{OksCamKalAkb:20} for details). This problem is caused by the fact that, since missing an object is typically more severe than producing a false alarm, object detectors produce more false alarm measurements than true measurements.

Since $\tilde{\beta}_{\mathrm{s},i}(a_{i} = j)$ in \eqref{eq:beta_enhanced} represents the likelihood that the legacy \ac{po} $i$ is associated to the measurement $j$, ideally $\mu_{i}(j) $ is large if \ac{po} $i$ is associated to the measurement $j$, and is equal to zero if they are not associated. Thus, we consider the following binary cross-entropy loss for object shape association, \vspace{0mm}i.e.,
\begin{align}
    \Set{L}_{\mathrm{a}} &= \frac{-1}{IJ} \sum_{i = 1}^{I} \sum_{j = 1}^{J}  \mu^{\text{gt}}_{i}(j) \ln \big( \sigma(\mu_{i}^{*}(j)) \big) \nn \\[0.8mm]
    &\hspace{17mm} + \big( 1 - \mu^{\text{gt}}_{i}(j) \big) \ln \big( 1 - \sigma(\mu_{i}^{*}(j)) \big) \vspace{0mm} \label{eq:loss_mu} \\[-2.5mm]
    \nn
\end{align}
where $\sigma(x) = 1/(1 + e^{-x})$ is the sigmoid function and $\mu^{\text{gt}}_{i} = [\mu^{\text{gt}}_{i}(1) \cdots \mu^{\text{gt}}_{i}(J)]^\T \in \{0,1\}^J$ is the pseudo ground truth association vector of legacy \ac{po} $i \rmv\in\rmv\{1,\dots,I\}$. In each pseudo ground truth association vector $\mu^{\text{gt}}_{i}$, at most one element is equal to one and all the other elements are equal to zero.  We apply $\mu_{i}^{*}(j)$ instead of $\mu_{i}(j)$ in the binary entropy loss \eqref{eq:loss_mu}. This is because the otherwise ReLU operation ``blocks'' certain gradients, i.e., gradients $\partial \Set{L}_{\mathrm{a}} / \partial \mu_{i}^{*}(j)$ are zero for negative values of $\mu_{i}^{*}(j)$. It was been observed, that by performing backpropagation by also making use of the gradients related to the negative values of $\mu_{i}^{*}(j)$, the \ac{gnn} can be trained more efficiently.

At each time step, pseudo ground truth association vectors are constructed from measurements and ground truth object tracks based on the following\vspace{.5mm} rules:
\begin{itemize}
    \item \textit{Get Measurement IDs:} First, the Euclidean distance between all ground truth positions and measurements is computed. Next, the Hungarian algorithm \cite{BarWilTia:B11} is performed to find the best association between ground truth positions and measurements. Finally, all measurements that have been associated with a ground truth position and have a distance to that ground truth position that is smaller than $T_{\text{dist}}$, inherit the ID of the ground truth position. All other measurements do not have an ID. \vspace{2mm}
    \item \textit{Update Legacy PO IDs:} Legacy POs inherit the ID from the previous time step. If a legacy PO with ID has a distance not larger than $T_{\text{dist}}$ to a ground truth position with the same ID, it keeps its ID. If a legacy PO $i \rmv\in\rmv \{1,\dots,I\}$ has the same ID as measurement $j \rmv\in\rmv \{1,\dots,J\}$, the entry $\mu^{\text{gt}}_{i}(j)$ is set to one. All other entries $\mu^{\text{gt}}_{i}(j)$, $i \rmv\in\rmv \{1,\dots,I\}$, $j \rmv\in\rmv \{1,\dots,J\}$ are set to zero. \vspace{2mm} 
     \item \textit{Introduce New PO IDs:} A new PO $j \rmv\in \{1,\dots,J\}$ inherits the ID from the corresponding measurement if the measurement has an ID that is different from the ID of any legacy \ac{po}. All other new POs do not have an\vspace{-.5mm} ID.
     \end{itemize}



\section{Numerical Results} \label{sec:exp}

To validate the performance of our method, we present results in an autonomous driving\vspace{-2mm} scenario. 

\subsection{Experimental Setup} \label{sec:experimentalSetup}

\subsubsection{Dataset}
\begin{figure}[!tbp]
    \centering
    \hspace*{-2.4mm}\scalebox{0.315}{\input{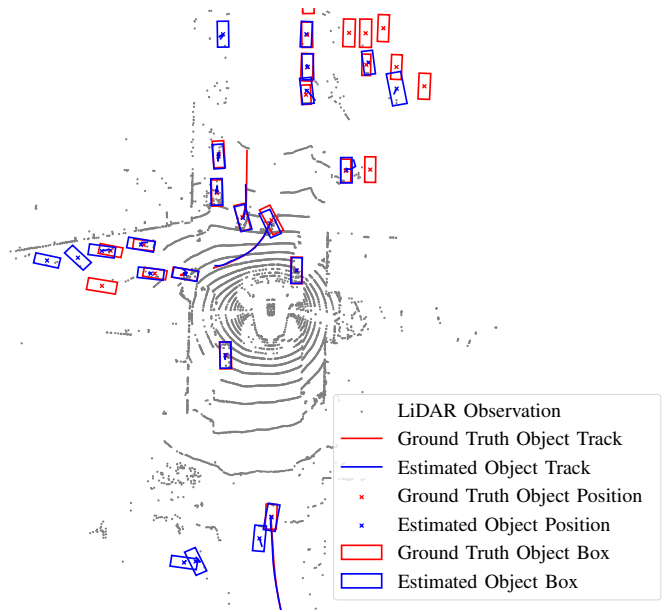}}
    \caption{Top-down view of the considered autonomous driving scene. For each ground truth and estimated vehicle, we plot tracks as well as positions and bounding boxes at the last time step. This scene is part of the \textit{nuScenes} autonomous driving dataset. }
    \label{fig:nusc-visualization}
   \vspace{-4mm}
\end{figure}

Our numerical evaluation is based on the \textit{nuScenes} autonomous driving dataset \cite{CaeBanLanVorLioXuKriPanBalBei:20}, which contains 1000 autonomous driving scenes. We use the official predefined dataset split, where 700 scenes are considered for training, 150 for validation, and 150 for testing. Each scene has a length of roughly 20 seconds and contains keyframes (frames with ground truth annotations) sampled at 2Hz. There are seven object classes.  The proposed \ac{mot} method and reference techniques are performed for each class individually. If not stated otherwise, all the operations described next are performed for each class separately. In this paper, we only consider LiDAR data provided by the nuScenes dataset. A scene of the considered autonomous driving application is shown in Fig. \ref{fig:nusc-visualization}.
\vspace{1mm}

\subsubsection{System Model}
The state of a \ac{po} $\V{x}_{k, n} \rmv\in\rmv \mathbb{R}^4$ consists of its 2-D position and 2-D velocity. Preprocessed measurements are extracted from the LiDAR data. For the extraction of preprocessed measurements, we employed the \textit{CenterPoint} \cite{YinZhoKra:21} detector which is based on deep learning\footnote{The measurements provided by the CenterPoint detector are further preprocessed using \ac{nms} with 3-D \ac{iou} \cite{NeuVan:06, PanLiWan:21} where the threshold is set to 0.1. }. Any preprocessed measurement $\V{z}_{k, j}$ consists of 2-D position, 2-D velocity, and a confidence score $0 < s_{k, j} \le 1$.

Object dynamics are modeled by a constant-velocity motion model \cite{ShaKirLi:B02}. Object tracking is performed in a global reference frame that is predefined for each scene \cite{CaeBanLanVorLioXuKriPanBalBei:20}. The considered \ac{roi} is defined by $[x_{\mathrm{e},k} - 54, x_{\mathrm{e},k} + 54] \times [y_{\mathrm{e},k}- 54, y_{\mathrm{e},k} + 54]$, where $(x_{\mathrm{e},k}, y_{\mathrm{e},k})$ is the 2-D position of the ``ego vehicle'' that is equipped with the LiDAR sensor. The \acp{pdf} that describe false alarms $f_{\text{FA}}(\cdot)$ and unknown objects $f_{\text{u}}(\cdot)$ are uniformly distributed over the \ac{roi}. The measurement model that defines the likelihood function $f(\V{z}_{k, j} | \V{x}_{k, n})$ is linear with additive Gaussian noise, i.e., $\V{z}_{k, j} = \M{H}_k\V{x}_{k, n} \rmv+\rmv \V{v}_{k, j} $, where $\V{v}_{k, j} \sim \Set{N}(\V{0}, \M{R})$ with $\M{R}$ being the diagonal covariance matrix. The probability of survival is set to $p_{\mathrm{s}} = 0.999$. The threshold for target declaration is $T_{\text{dec}} = 0.5$. 

The pruning threshold discussed in Section~\ref{sec:estimationDetection} is set to $T_{\text{pru}} = 10^{-3}$. In addition, we also prune new POs with $\tilde{p}(b_{k, j} = 0) < 0.8$ to further reduce the number of false objects and computational complexity. All other parameters used in the system model are extracted from training data. For the \ac{bp} part of the proposed \ac{nebp} method, we use the particle-based implementation introduced in \cite{MeyBraWilHla:J17}.
\vspace{1mm}

\subsubsection{Performance Metrics}
We consider using the \ac{amota} metric proposed in \cite{WenWanHelKit:20} to evaluate the performance of our algorithm.  In addition, we also use the widely used CLEAR metrics \cite{BerSti:08} and track quality measures \cite{WuNev:06} that include the number of \ac{ids} and track \ac{frag}. The number of \acp{ids} is increased if a ground truth object is matched to an estimated object with index $i$ at the current time step, while it was matched to an estimated object with index  $j \ne i$ at a previous time. The number of \acp{frag} is increased if a ground truth object is matched to an estimated object at the previous time step, but it is not matched to any estimated objects at the current time step. Note that \ac{amota} is the primary metric used by the nuScenes tracking challenge\vspace{-3mm} \cite{CaeBanLanVorLioXuKriPanBalBei:20}.

\subsection{Implementation Details} \label{subsec:exp_imp}

For shape features extraction as discussed in Section \ref{sec:feat_extract}, a neural network $g_{\text{shape}}(\cdot) \triangleq g_{\text{shape}, 1}(\cdot) = g_{\text{shape}, 2}(\cdot)$ that consists of two stages is introduced. The first stage is a \textit{VoxelNet} \cite{ZhuTuz:18}, a neural network architecture that is used as the backbone for a variety of object detectors \cite{YinZhoKra:21,ZhuJiaZhoLiYu:19}. The VoxelNet takes the LiDAR scan $\Set{Z}_k$ as input, and outputs a 3D tensor of size $180 \times 180 \times 512$. This tensor is typically referred to as feature map. The first two dimensions of the feature map form a grid with $180 \times 180$ elements that cover the \ac{roi}. For each grid point, there is a feature vector with $512$ elements. The second stage is a \ac{cnn} that consists of two convolutional layers and a single-hidden-layer \ac{mlp}. Here, we use a \ac{cnn} since it has fewer trainable parameters compared to an MLP and is thus easier to train. Note that \acp{cnn} have been widely used for feature extraction \cite{LecBosDenHenHowHubJac:89, ZhuTuz:18}. The \ac{cnn} extracts shape features from a reduced feature map, as discussed next.

For the extraction of shape features in the second stage, at first, the grid point of the feature map that corresponds to the considered \acp{po} or measurements is located. Then, the feature vector at this grid point and the 8 feature vectors at adjacent grid points are extracted. As a result, for each \ac{po} and each measurement, a reduced feature map of size $3 \times 3 \times 512$ is extracted. This reduced feature map is then used as the input of a \ac{cnn}. Finally, the \ac{cnn} computes the shape feature $\V{h}_{a_i, \text{shape}}$ or $\V{h}_{b_j, \text{shape}}$. The considered feature map of dimension $180 \times 180 \times 512$ has been precomputed by the CenterPoint \cite{YinZhoKra:21} method. The same VoxelNet is shared across all seven object classes. Its parameters remain fixed during the training of the proposed method.

The other neural networks $g_{\text{e}}(\cdot)$, $g_{\text{n}}(\cdot)$, $g_{\text{d}}(\cdot)$, $g_{\mathrm{s}}(\cdot)$, and $g_{\text{motion}}(\cdot) \rmv\triangleq\rmv g_{\text{motion}, 1}(\cdot) \rmv=\rmv g_{\text{motion}, 2}(\cdot)$ are \acp{mlp} with a single hidden layer and a leaky ReLU activation function. All feature vectors, i.e., $\V{h}_{a_i, \text{motion}}$ and $\V{h}_{a_i, \text{shape}}$, $i \in \{1,\dots,I\}$ as well as $\V{h}_{b_j, \text{motion}}$ and $\V{h}_{b_j, \text{shape}}$, $j \in \{1,\dots,J\}$, consist of 128 elements. The number of \ac{gnn} iterations is $P = 3$. Training of the proposed method is performed based on the \textit{Adam optimizer} \cite{KinBa:14}. The batch size, learning rate, and the number of ``epochs'', i.e., the number of times the Adam optimizer processes the entire training dataset, are set to $1$, $10^{-4}$, and $8$, respectively. The hyperparameter $\epsilon$ in \eqref{eq:loss_omega} is set to 0.1 and the threshold $T_{\text{dist}}$ for the pseudo ground truth extraction discussed in Section \ref{sec:loss} is\vspace{0mm} set to 2 meters. 

Evaluation of \ac{amota} requires a score for each estimated object. It was observed that a high \ac{amota} performance is obtained by calculating the estimated object score as a combination of existence probability and measurement score. In particular, for legacy \ac{po} $i \rmv\in\rmv \{1,\dots,I\}$ we calculate an estimated object score\vspace{-.5mm} as
\begin{equation}
 \label{eq:estimatedObjectScore}
 \underline{s}{}_i = \tilde{p}(\underline{r}{}_i = 1) + \sum_{j = 1}^{J} \tilde{p}(a_i = j) s_j.
 \vspace{-.3mm}
\end{equation}
For new PO $j \rmv\in\rmv \{1,\dots,J\}$ the estimated object score is given by\vspace{-2mm} $\overline{s}{}_j = \tilde{p}(\overline{r}{}_j = 1) + s_j $.

\subsection{Calibration}
\label{sec:calibration}
The calibration of the sigmoid introduced in \eqref{eq:nebp_omega} is performed as follows. For training, we set $T\rmv=\rmv1$ and $\delta \rmv=\rmv 0$. However, for inference we set $T\rmv>\rmv0$ and $\delta\rmv>\rmv0$ such that the sigmoid in \eqref{eq:nebp_omega} transitions to one quicker and for smaller values of $\omega^{*}_j$ . The different calibration values for inference are necessary because the loss function used for training and the AMOTA metric used for performance evaluation behave differently. In particular, for performance evaluation based on AMOTA, missing an object is significantly more severe than a false object. (Note that the AMOTA metric can not directly be used for training because it is not differentiable~\cite{Bishop:B06}.) The calibration values $T$ and $\delta$ used for inference are selected based on a grid search over possible values  $T \in \Set{T} = \{0.5, 1, 2, 4, 8, 16, 32, +\infty\}$ and $\delta \in \Set{\Delta} = \big\{ \delta \ist | \ist \sigma(\delta) \in \{ 0.01, 0.02, 0.05, 0.1, 0.2\} \big\}$. The values for $\delta$ and $T$ that achieved the highest \ac{amota} value on a subset of the training set were obtained as $\sigma^{-1}(0.2, 0.02, 0.2, 0.02, 0.2, 0.01, 0.01)$ and $(\infty, 8, 1, \infty, 1, 8, 0.5)$ for the classes bicycle, bus, car, motorcycle, pedestrian, trailer, and truck, respectively.



\subsection{Performance Evaluation} \label{subsec:performance}

For performance evaluation, we use state-of-the-art reference methods that all use measurements provided by the CenterPoint detector \cite{YinZhoKra:21}, which was the best LiDAR-only object detector for the nuScenes dataset at the time of the submission of this paper. In particular, \ac{bp} refers to the conventional \ac{bp}-based \ac{mot} method \cite{MeyKroWilLauHlaBraWin:J18}. CenterPointT refers to the tracking method proposed in \cite{YinZhoKra:21}. It uses a heuristic to create new tracks and a greedy matching algorithm based on the Euclidean distance to associate measurements provided by the CenterPoint detector. The methods in \cite{BenSchZel:21,ChiLiAmbBoh:21, PanLiWan:21} all follow a similar strategy. The CBMOT method \cite{BenSchZel:21} adopts a score update function for estimated object scores. Chiu et al. \cite{ChiLiAmbBoh:21} make use of a hybrid distance that combines the Mahalanobis distance with a proposed deep feature distance. SimpleTrack \cite{PanLiWan:21} uses the \ac{giou} as the distance for measurement association. In SimpleTrack \cite{PanLiWan:21}, the object detector is also applied to non-keyframes and has a measurement rate of 10Hz. The Immortal tracker \cite{WanChePanWanZha:21} has a measurement rate of 2Hz. It follows the tracking approach of SimpleTrack, except that it never terminates tracks. PMB \cite{LiuBaiXiaHuaZhu:22} implements a Poisson multi-Bernoulli filter for \ac{mot} that relies on a global nearest neighbor approach for data association. Finally, OGR3MOT \cite{ZaeDaiLinDanVan:22} utilizes a network flow formulation and transforms the data association problem into a classification problem.

\begin{table}[!tbp]
    \centering
    \caption{Performance results: nuScenes test set}
    \label{tab:nusc-test}
    \begin{tabular}{c|c|ccc}
    \hline
    Method                            & Modalities   & AMOTA $\uparrow$ & IDS $\downarrow$ & Frag $\downarrow$ \\ \hline \hline
    CenterPointT \cite{YinZhoKra:21}   & LiDAR        & 0.638            & 760              & 529               \\
    CBMOT \cite{BenSchZel:21}         & LiDAR        & 0.649            & 557              & 450               \\
    Chiu et al. \cite{ChiLiAmbBoh:21} & LiDAR+Camera & 0.655            & 1043             & 717               \\
    OGR3MOT \cite{ZaeDaiLinDanVan:22} & LiDAR        & 0.656            & 288              & 371               \\
    SimpleTrack \cite{PanLiWan:21}    & LiDAR        & 0.668            & 575              & 591               \\
    Immortal \cite{WanChePanWanZha:21}    & LiDAR        & 0.677            & 320              & 477               \\ 
    PMB \cite{LiuBaiXiaHuaZhu:22} & LiDAR        & 0.678            & 770              & 431 \\ \hline
    BP                                & LiDAR        & 0.666            & \textbf{182}     & \textbf{245}               \\
    NEBP (proposed)                   & LiDAR        & \textbf{0.683}   & 227              & 299      \\ \hline
    \end{tabular}
    \vspace{-2mm}
\end{table}

\begin{table}[!tbp]
    \centering
    \caption{Performance results: nuScenes validation set}
    \label{tab:nusc-val}
    \begin{tabular}{c|c|ccc}
    \hline
    Method                           & Modalities   & AMOTA $\uparrow$ & IDS $\downarrow$ & Frag $\downarrow$ \\ \hline \hline
    CenterPointT \cite{YinZhoKra:21}   & LiDAR        & 0.665            & 562              & 424               \\
    CBMOT \cite{BenSchZel:21}         & LiDAR        & 0.675            & 494              & -                 \\
    Chiu et al. \cite{ChiLiAmbBoh:21} & LiDAR+Camera & 0.687            & -                & -                 \\
    OGR3MOT \cite{ZaeDaiLinDanVan:22} & LiDAR        & 0.693            & 262              & 332               \\
    SimpleTrack \cite{PanLiWan:21}    & LiDAR        & 0.696            & 405              & -                 \\
    Immortal \cite{WanChePanWanZha:21}    & LiDAR        & 0.702            & 385              & -                 \\
    PMB \cite{LiuBaiXiaHuaZhu:22} & LiDAR        & 0.707            & 650              & 345 \\ \hline
    BP                                & LiDAR        & 0.698            & \textbf{161}     & \textbf{250}      \\
    NEBP (proposed)                   & LiDAR        & \textbf{0.708}   & 172              & 271               \\ \hline
    \end{tabular}
  \vspace{-2mm}
\end{table}

\begin{figure*}[!t]
    \centering
    \subfloat[]{
        \includegraphics[width=0.24\linewidth]{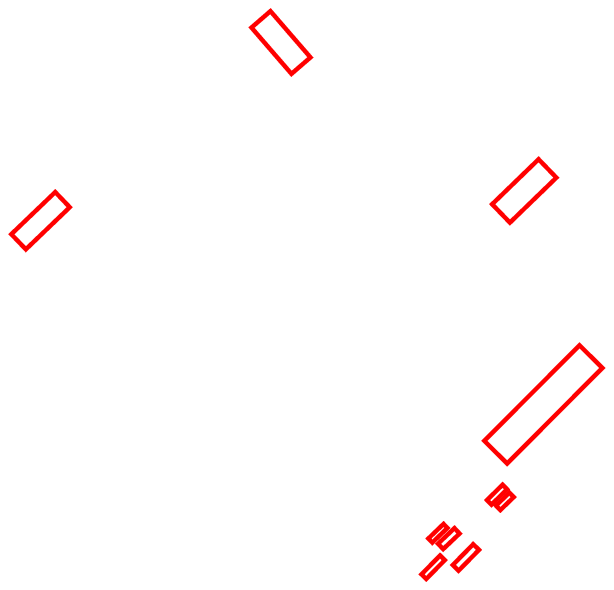}
        \label{subfig:gt}}
    \subfloat[]{
        \includegraphics[width=0.24\linewidth]{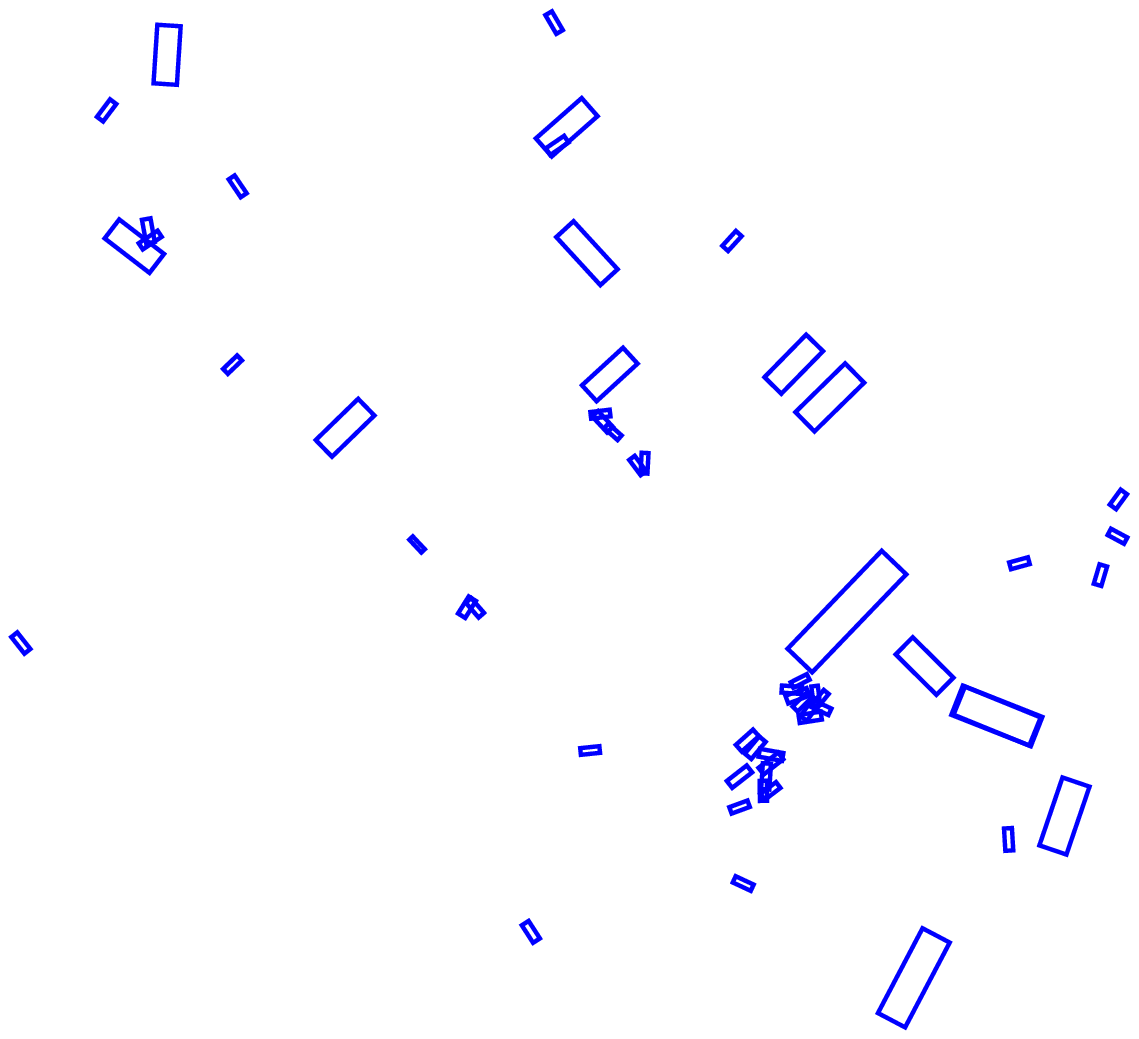} 
        \label{subfig:centerPoint}}
    \subfloat[]{
        \includegraphics[width=0.24\linewidth]{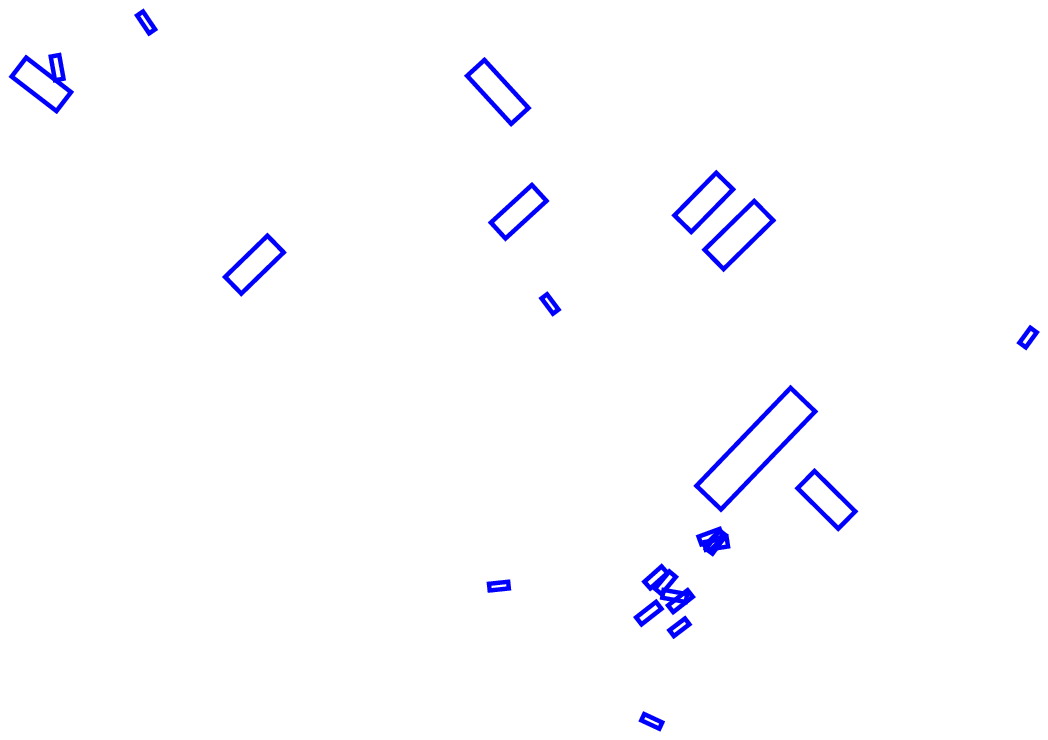} 
        \label{subfig:bp}}
    \subfloat[]{
        \includegraphics[width=0.24\linewidth]{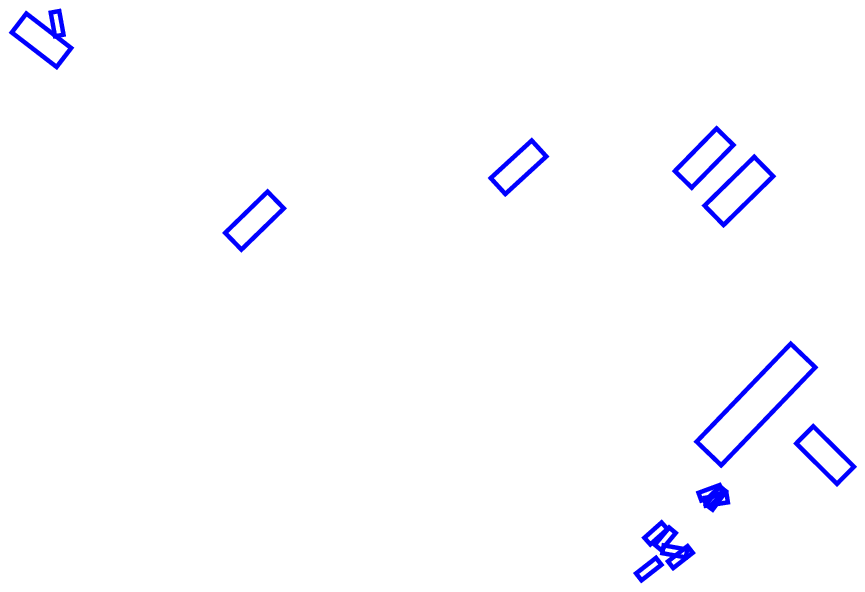} 
        \label{subfig:nebp}}
    
    \captionsetup{singlelinecheck = false, justification=justified}	
    \caption{Top-down view and single time step for an example autonomous driving scene.  Ground truth objects~(a) as well as estimated object provided by CenterPointT~(b), \ac{bp}~(c), and \ac{nebp}~(d) are shown. Note that \ac{bp} and \ac{nebp} do not provide object size and orientation estimates. Thus, for each estimated object, the size and orientation of the measurement with the largest association probability are shown. }
    \vspace{-1mm}
    \label{fig:qualitative_result}
\end{figure*}

\begin{table}[!tbp]
    \centering
    \caption{Results for different detectors: nuScenes validation set}
    \label{tab:ablation-detector}
    \begin{tabular}{c|c|ccc}
    \hline
    Detector                                                  & Method          & AMOTA $\uparrow$ & IDS $\downarrow$ & Frag $\downarrow$ \\ \hline \hline
    \multirow{2}{*}{PointPillar \cite{LanVorCaeZhoYanBei:19}} & BP              & 0.360            & \textbf{226}              & 658               \\
                                                              & NEBP (proposed) & \textbf{0.398}   & 270     & \textbf{418}      \\ \hline
    \multirow{2}{*}{Megvii \cite{ZhuJiaZhoLiYu:19}}           & BP              & 0.637            & \textbf{108}     & 262               \\
                                                              & NEBP (proposed) & \textbf{0.651}   & 136      & \textbf{252}      \\ \hline
    \multirow{2}{*}{CenterPoint \cite{YinZhoKra:21}}          & BP              & 0.698            & \textbf{161}      & \textbf{250}      \\
                                                              & NEBP (proposed) & \textbf{0.708}   & 172              & 271               \\ \hline
    \end{tabular}
    \vspace{-2mm}
\end{table}

In Table~\ref{tab:nusc-test} and \ref{tab:nusc-val}, we present the tracking performance of the considered methods on the nuScenes validation and test sets based on measurements provided by the CenterPoint detector. The symbol ``-'' in Table \ref{tab:nusc-val} indicates that the metric is not reported. It can be seen that the proposed \ac{nebp} approach outperforms all reference methods in terms of \ac{amota} performance. Furthermore, it can be observed, that \ac{bp} and \ac{nebp} achieve a much lower \ac{ids} and \ac{frag} metric compared to the reference methods. This is because both \ac{bp} and \ac{nebp} make use of a statistical model to determine the initialization and termination of tracks \cite{MeyKroWilLauHlaBraWin:J18}, which is more robust compared to the heuristic track management performed by other reference methods. The improved \ac{amota} performance of \ac{nebp} over \ac{bp} comes at the cost of a slightly increased \ac{ids} and \ac{frag}. Qualitative results for a single time step of an example autonomous driving scene are shown in Fig. \ref{fig:qualitative_result}. It can be seen that compared to CenterPointT, \ac{bp} can reduce the number of false objects significantly, while \ac{nebp} can reduce the number of false alarms even further. We also report estimation performance results based on the \ac{gospa} metric \cite{RahGarSve:17}. This metric can be split up into three components, namely, localization error, false estimated objects, and missed ground truth objects. \ac{nebp} outperforms \ac{bp} in all three components.

\begin{table}[!tbp]
\vspace{2mm}
    \centering
    \caption{GOSPA  results: nuScenes validation set}
    \label{tab:gospa}
    \begin{tabular}{c|c|ccc}
    \hline 
    Method          & GOSPA $\downarrow$ & Localization $\downarrow$  & False $\downarrow$ & Missed $\downarrow$   \\ \hline \hline
    BP              & 1.163              & 0.327                      & 0.127              & 0.709          \\
    NEBP (proposed) & \textbf{1.140}     & \textbf{0.318}             & \textbf{0.123}     & \textbf{0.699} \\ \hline
    \end{tabular}
    \vspace{-2mm}
\end{table}

We also compare the performance of the proposed NEBP method with BP based on measurements provided by different object detectors. In particular, in addition to measurements provided by the CenterPoint \cite{YinZhoKra:21} detector, we also consider measurements provided by the PointPillar \cite{LanVorCaeZhoYanBei:19} and the Megvii \cite{ZhuJiaZhoLiYu:19} detectors. Results based on the nuScenes validation set are shown in Table \ref{tab:ablation-detector}. For all three detectors, \ac{nebp} can outperform \ac{bp} in terms of \ac{amota}, and at the same time maintain a similar number of \ac{ids} and \ac{frag}. These results indicated that the proposed \ac{nebp} method is robust with respect to the chosen object detector.

All experiments were executed on a single Nvidia P100 GPU. For training, eight epochs of the nuScenes training set were performed. The total training time was measured as 30 hours. The inference times of \ac{bp} and \ac{nebp} applied to the nuScenes validation were measured as 658 seconds and 1137 seconds. These times do not include the execution of the object detector, which yields a runtime of 822 seconds. \ac{nebp} has a higher computational complexity compared to \ac{bp} due to the additional operations discussed in Section \ref{subsec:complexity-analysis}.



\subsection{Ablation Study}
\label{sec:ablaStud}
In this section, we analyze the contribution of different algorithmic components to the overall performance of our \ac{nebp} method. In particular, we analyze the degradations of \ac{nebp} performance that are the result of the ablation of specific algorithmic components.
All ablation studies are based on the nuScenes validation set, and measurements provided by the CenterPoint detector \cite{YinZhoKra:21}.

We aim to quantify the performance benefits of \ac{nebp} resulting from using shape features as an additional input. We consider three \ac{nebp} variants: ``NEBP-m'',  ``NEBP-r'' and ``NEBP-a''. NEBP-m does not make use of shape features. In particular, it initializes the \ac{gnn} node embeddings only based on motion features, i.e., $\V{h}_{a_i}^{(1)} \rmv=\rmv [\V{h}_{a_i, \text{motion}}^\T$ $\V{0}^\T]^\T$ and $\V{h}_{b_j}^{(1)} \rmv=\rmv [\V{h}_{b_j, \text{motion}}^\T$ $\V{0}^\T]^\T\rmv$. The two variants NEBP-r and NEBP-a are introduced to quantify the benefits of false alarm rejection and object shape association, respectively. More specifically, NEBP-r focuses on false alarm rejection only, i.e., in \eqref{eq:nebp_combine_a} object shape association is deactivated by setting $\mu_i(j) \rmv=\rmv 0, i \in \{1,\dots,I\}, j \in \{1,\dots,J\}$. Similarly, NEBP-a focuses on object shape association only, i.e., in \eqref{eq:nebp_combine_a} and \eqref{eq:nebp_combine_b} false alarm rejection is deactivated by setting  $\omega_j = 1, j \in \{1,\dots,J\}$. 

\begin{table}[!tbp]
    \centering
    \caption{Ablation studies: nuScenes validation set}
    \label{tab:nusc-ablation-ra}
    \begin{tabular}{c|ccc}
    \hline
    Method  & AMOTA $\uparrow$ & IDS $\downarrow$ & Frag $\downarrow$ \\ \hline \hline
    BP      & 0.698            & \textbf{161}     & 250      \\
    NEBP-m  & 0.698            & 183              & 228               \\
    NEBP-a  & 0.703            & 195              & 259               \\
    NEBP-r  & 0.706            & 178              & \textbf{227}               \\
    NEBP-nc & 0.703            & 180              & 250               \\
    NEBP (proposed)    & \textbf{0.708}   & 172              & 271               \\ \hline
    \end{tabular}
      \vspace{0mm}
\end{table}

Table \ref{tab:nusc-ablation-ra} shows the tracking performance of these \ac{nebp} variants, conventional \ac{bp}, and the proposed \ac{nebp} method. It can be seen that NEBP-m can not achieve any performance improvements compared to \ac{bp}. This is not surprising since object motion is already modeled accurately by the statistical model and, compared to \ac{bp}, NEBP-m does not make use of any additional information. On the other hand, both NEBP-a and NEBP-r can achieve an improved \ac{amota} performance compared to \ac{bp}. This is because NEBP-a and NEBP-r incorporate additional information in the form of shape features and address the fact that the statistical model used by \ac{bp} does not accurately model the true data-generating process. In particular, in the statistical model, false alarm measurements are uniformly distributed over the \ac{roi}.
Furthermore, false alarm measurement and their number are also assumed to be independent and identically distributed across time. However, these assumptions do often not hold in real-world \ac{mot} applications such as the considered autonomous driving scenario. This is because physical structures and other reflecting features in the environment can generate so-called persistent false alarm measurements. These false alarm measurements are not uniformly distributed and are not independent across time. Thus, they are not accurately represented by the considered statistical model. This model mismatch degrades tracking performance and is addressed by false alarm rejection performed by NEBP and NEBP-r. Object shape association as performed by NEBP and NEBP-a improves data association by using object shape information provided by shape features.

Finally, Table \ref{tab:nusc-ablation-ra} also reports performance improvements that result from the calibration process discussed in Section \ref{sec:calibration}. In particular, ``NEBP-nc'' is the NEBP variant where no calibration has been performed, i.e., we have $T=1$ and $\delta = 0$ for training and inference. It can be seen that calibration can significantly improve the performance of \ac{nebp}. 

The effect of temperature parameters $T\rmv>\rmv0$ for five representative object classes is shown in Table \ref{tab:nusc-temperature}.  For each class, the bias $\delta$ is fixed to the value provided above. Note that $T = +\infty$, is equivalent to discarding all measurements with $\omega_j < \sigma(\delta)$. It can be seen that \ac{nebp} does not always achieve the best \ac{amota} for $T = +\infty$. In cases where it is difficult to determine whether a measurement is a false alarm, using a temperature $T < +\infty$ can be more robust, as it does not directly discard the measurements with $\omega_j < \sigma(\delta)$, but instead reduces the estimated object score \eqref{eq:estimatedObjectScore} of \acp{po} that are likely to generate these\vspace{1.5mm} measurements.

\begin{table}[!tbp]
    \centering
    \caption{AMOTA results for different sigmoid temperatures: nuScenes validation set}
    \label{tab:nusc-temperature}
    \begin{tabular}{l|ccccc}
    \hline
    Temperature        & bicycle        & pedestrian     & motorcycle     & trailer        & truck          \\ \hline
    $T = 0.5$     & 0.528          & 0.807          & 0.737          & 0.568          & 0.662          \\
    $T = 1$       & 0.534          & 0.809          & 0.727          & 0.565          & 0.663          \\
    $T = 2$       & 0.527          & 0.810          & 0.716          & 0.563          & 0.662          \\
    $T = 4$       & \textbf{0.550} & \textbf{0.815} & 0.727          & 0.566          & \textbf{0.664} \\
    $T = 8$       & 0.544          & 0.814          & \textbf{0.739} & \textbf{0.569} & 0.662          \\
    $T = 16$      & 0.540          & 0.813          & 0.730          & 0.562          & 0.662          \\
    $T = 32$      & 0.547          & 0.814          & 0.736          & 0.567          & 0.657          \\
    $T = +\infty$ & \textbf{0.550} & 0.813          & \textbf{0.739} & 0.559          & 0.656          \\ \hline
    \end{tabular}
    \vspace{0mm}
\end{table}

\acresetall
\section{Conclusion}
\label{sec:conclusion}
In this paper, we present a \ac{nebp} method for \ac{mot} that enhances the solution of model-based \ac{bp} by making use of shape features learned from raw sensor data. Our approach conjectures that learned information can reduce model mismatch and thus improve data association and rejection of false alarms. A \ac{gnn} that matches the topology of the factor graph used for model-based data association is introduced. For false alarm rejection, the \ac{gnn} identifies measurements that are likely false alarms. For object shape association, the \ac{gnn} computes corrections terms that result in more accurate association probabilities. The proposed approach can improve the object declaration and state estimation performance of \ac{bp} while preserving its favorable scalability of the computational complexity. Furthermore, the proposed \ac{nebp} method inherits the robust track management of \ac{bp}-based algorithms. We employed the nuScenes autonomous driving dataset for performance evaluation and demonstrated state-of-the-art object tracking performance. Due to robust track management \ac{nebp} yields a much lower number of \ac{ids} and track \ac{frag} compared to non-\ac{bp}-based reference\vspace{0mm} methods. A promising direction for future research is an application of the proposed \ac{nebp} approach to multipath-aided localization \vspace{0mm} \cite{LeiMeyHlaWitTufWin:J19}.


.





\ifCLASSOPTIONcaptionsoff
  \newpage
\fi


\renewcommand{\baselinestretch}{0.985}
\bibliographystyle{IEEEtran}
\bibliography{IEEEabrv,StringDefinitions,Books,Papers,ref}

\begin{thebibliography}{10}
\providecommand{\url}[1]{#1}
\csname url@samestyle\endcsname
\providecommand{\newblock}{\relax}
\providecommand{\bibinfo}[2]{#2}
\providecommand{\BIBentrySTDinterwordspacing}{\spaceskip=0pt\relax}
\providecommand{\BIBentryALTinterwordstretchfactor}{4}
\providecommand{\BIBentryALTinterwordspacing}{\spaceskip=\fontdimen2\font plus
\BIBentryALTinterwordstretchfactor\fontdimen3\font minus
  \fontdimen4\font\relax}
\providecommand{\BIBforeignlanguage}[2]{{%
\expandafter\ifx\csname l@#1\endcsname\relax
\typeout{** WARNING: IEEEtran.bst: No hyphenation pattern has been}%
\typeout{** loaded for the language `#1'. Using the pattern for}%
\typeout{** the default language instead.}%
\else
\language=\csname l@#1\endcsname
\fi
#2}}
\providecommand{\BIBdecl}{\relax}
\BIBdecl

\bibitem{BarWilTia:B11}
Y.~Bar-Shalom, P.~K. Willett, and X.~Tian, \emph{{Tracking and Data Fusion: A
  Handbook of Algorithms}}.\hskip 1em plus 0.5em minus 0.4em\relax Storrs, CT:
  Yaakov Bar-Shalom, 2011.

\bibitem{Mah:B07}
R.~Mahler, \emph{{Statistical Multisource-Multitarget Information
  Fusion}}.\hskip 1em plus 0.5em minus 0.4em\relax Norwood, MA: Artech House,
  2007.

\bibitem{MeyBraWilHla:J17}
F.~Meyer, P.~Braca, P.~Willett, and F.~Hlawatsch, ``{A scalable algorithm for
  tracking an unknown number of targets using multiple sensors},'' \emph{{IEEE}
  Trans. Signal Process.}, vol.~65, no.~13, pp. 3478--3493, Jul. 2017.

\bibitem{MeyKroWilLauHlaBraWin:J18}
F.~Meyer, T.~Kropfreiter, J.~L. Williams, R.~A. Lau, F.~Hlawatsch, P.~Braca,
  and M.~Z. Win, ``Message passing algorithms for scalable multitarget
  tracking,'' \emph{Proc. {IEEE}}, vol. 106, no.~2, pp. 221--259, Feb. 2018.

\bibitem{MeyWil:J21}
F.~Meyer and J.~L. Williams, ``Scalable detection and tracking of geometric
  extended objects,'' \emph{{IEEE} Trans. Signal Process.}, vol.~69, pp.
  6283--6298, 2021.

\bibitem{Wil:J15}
J.~L. Williams, ``{Marginal multi-Bernoulli filters: RFS derivation of MHT,
  JIPDA and association-based MeMBer},'' \emph{{IEEE} Trans. Aerosp. Electron.
  Syst.}, vol.~51, no.~3, pp. 1664--1687, Jul. 2015.

\bibitem{GarWilGraSve:J18}
{\'A}.~F. Garc{\'\i}a-Fern{\'a}ndez, J.~L. Williams, K.~Granstr{\"o}m, and
  L.~Svensson, ``Poisson multi-{Bernoulli} mixture filter: Direct derivation
  and implementation,'' \emph{{IEEE} Trans. Aerosp. Electron. Syst.}, vol.~54,
  no.~4, pp. 1883--1901, Feb. 2018.

\bibitem{SchBenRosKriGra:18}
S.~Scheidegger, J.~Benjaminsson, E.~Rosenberg, A.~Krishnan, and K.~Granström,
  ``Mono-camera {3D} multi-object tracking using deep learning detections and
  {PMBM} filtering,'' in \emph{Proc. IEEE IV-18}, Jun. 2018, pp. 433--440.

\bibitem{SolMeyBraHla:J19}
G.~Soldi, F.~Meyer, P.~Braca, and F.~Hlawatsch, ``Self-tuning algorithms for
  multisensor-multitarget tracking using belief propagation,'' \emph{{IEEE}
  Trans. Signal Process.}, vol.~67, no.~15, pp. 3922--3937, Aug. 2019.

\bibitem{PanMorRad:21}
S.~Pang, D.~Morris, and H.~Radha, ``{3D} multi-object tracking using random
  finite set-based multiple measurement models filtering {(RFS-M$^3$)} for
  autonomous vehicles,'' in \emph{Proc. ICRA-21}, Jun. 2021, pp.
  13\,701--13\,707.

\bibitem{LiuBaiXiaHuaZhu:22}
J.~Liu, L.~Bai, Y.~Xia, T.~Huang, and B.~Zhu, ``{GNN-PMB}: A simple but
  effective online {3D} multi-object tracker without bells and whistles,''
  \emph{arXiv preprint arxiv.2206.10255}, 2022.

\bibitem{ZhaMey:C21}
W.~Zhang and F.~Meyer, ``Graph-based multiobject tracking with embedded
  particle flow,'' in \emph{Proc. IEEE RadarConf-21}, May 2021, pp. 1--6.

\bibitem{YinZhoKra:21}
T.~Yin, X.~Zhou, and P.~Krahenbuhl, ``Center-based {3D} object detection and
  tracking,'' in \emph{Proc. CVPR-21}, Jun. 2021, pp. 11\,784--11\,793.

\bibitem{MeyWin:J20}
F.~{Meyer} and M.~Z. {Win}, ``Scalable data association for extended object
  tracking,'' \emph{IEEE Trans. Signal Inf. Process. Netw.}, vol.~6, pp.
  491--507, May 2020.

\bibitem{ChiPriLiBoh:20}
H.-k. Chiu, A.~Prioletti, J.~Li, and J.~Bohg, ``Probabilistic {3D} multi-object
  tracking for autonomous driving,'' \emph{arXiv preprint arXiv:2001.05673},
  2020.

\bibitem{WenWanHelKit:20}
X.~Weng, J.~Wang, D.~Held, and K.~Kitani, ``{3D} multi-object tracking: A
  baseline and new evaluation metrics,'' in \emph{Proc. IROS-20}, Oct. 2020,
  pp. 10\,359--10\,366.

\bibitem{ZaeDaiLinDanVan:22}
J.-N. Zaech, D.~Dai, A.~Liniger, M.~Danelljan, and L.~Van~Gool, ``Learnable
  online graph representations for {3D} multi-object tracking,'' \emph{{IEEE}
  Robot. Autom. Lett.}, vol.~7, no.~2, pp. 5103--5110, Jan. 2022.

\bibitem{RanMahGebMhaRamTri:21}
A.~Rangesh, P.~Maheshwari, M.~Gebre, S.~Mhatre, V.~Ramezani, and M.~M. Trivedi,
  ``{TrackMPNN}: A message passing graph neural architecture for multi-object
  tracking,'' \emph{arXiv preprint arXiv:2101.04206}, 2021.

\bibitem{PanLiWan:21}
Z.~Pang, Z.~Li, and N.~Wang, ``{SimpleTrack}: Understanding and rethinking {3D}
  multi-object tracking,'' \emph{arXiv preprint arXiv:2111.09621}, 2021.

\bibitem{WanChePanWanZha:21}
Q.~Wang, Y.~Chen, Z.~Pang, N.~Wang, and Z.~Zhang, ``{Immortal Tracker}:
  Tracklet never dies,'' \emph{arXiv preprint arXiv:2111.13672}, 2021.

\bibitem{WenWanManKit:20}
X.~Weng, Y.~Wang, Y.~Man, and K.~M. Kitani, ``{GNN3DMOT}: Graph neural network
  for {3D} multi-object tracking with {2D-3D} multi-feature learning,'' in
  \emph{Proc. CVPR-20}, Jun. 2020, pp. 6499--6508.

\bibitem{ChiLiAmbBoh:21}
H.-k. Chiu, J.~Li, R.~Ambrus, and J.~Bohg, ``Probabilistic {3D} multi-modal,
  multi-object tracking for autonomous driving,'' in \emph{Proc. ICRA-21}, Jun.
  2021, pp. 14\,227--14\,233.

\bibitem{LiLeiVen:J22}
X.~Li, E.~Leitinger, A.~Venus, and F.~Tufvesson, ``Sequential detection and
  estimation of multipath channel parameters using belief propagation,''
  \emph{{IEEE} Trans. Wireless Commun.}, vol.~21, no.~10, pp. 8385--8402, 2022.

\bibitem{ZhuTuz:18}
Y.~Zhou and O.~Tuzel, ``Voxelnet: End-to-end learning for point cloud based
  {3D} object detection,'' in \emph{Proc. CVPR-18}, Jun. 2018, pp. 4490--4499.

\bibitem{LanVorCaeZhoYanBei:19}
A.~H. Lang, S.~Vora, H.~Caesar, L.~Zhou, J.~Yang, and O.~Beijbom,
  ``{PointPillars}: Fast encoders for object detection from point clouds,'' in
  \emph{Proc. CVPR-19}, Jun. 2019, pp. 12\,697--12\,705.

\bibitem{ZhuJiaZhoLiYu:19}
B.~Zhu, Z.~Jiang, X.~Zhou, Z.~Li, and G.~Yu, ``Class-balanced grouping and
  sampling for point cloud {3D} object detection,'' \emph{arXiv:1908.09492},
  2019.

\bibitem{RenHeGirSun:15}
S.~Ren, K.~He, R.~Girshick, and J.~Sun, ``Faster {R-CNN}: Towards real-time
  object detection with region proposal networks,'' in \emph{Proc. NeurIPS-15},
  vol.~28, Dec. 2015, pp. 91--99.

\bibitem{ShiWanLi:19}
S.~Shi, X.~Wang, and H.~Li, ``{PointRCNN}: {3D} object proposal generation and
  detection from point cloud,'' in \emph{Proc. CVPR-19}, Jun. 2019, pp.
  770--779.

\bibitem{SimBulPorLop:19}
A.~Simonelli, S.~R. Bulo, L.~Porzi, M.~L{\'o}pez-Antequera, and
  P.~Kontschieder, ``Disentangling monocular {3D} object detection,'' in
  \emph{Proc. CVPR-19}, Jun. 2019, pp. 1991--1999.

\bibitem{OkuTalFreLitLow:04}
K.~Okuma, A.~Taleghani, N.~d. Freitas, J.~J. Little, and D.~G. Lowe, ``A
  boosted particle filter: Multitarget detection and tracking,'' in \emph{Proc.
  ECCV-04}, 2004, pp. 28--39.

\bibitem{BreReiLeiKolVan:10}
M.~D. Breitenstein, F.~Reichlin, B.~Leibe, E.~Koller-Meier, and L.~Van~Gool,
  ``Online multiperson tracking-by-detection from a single, uncalibrated
  camera,'' \emph{{IEEE} Trans. Pattern Anal. Mach. Intell.}, vol.~33, no.~9,
  pp. 1820--1833, 2010.

\bibitem{Kuh:55}
H.~W. Kuhn, ``The {Hungarian} method for the assignment problem,'' \emph{NRL
  quarterly}, vol.~2, no. 1-2, pp. 83--97, Mar. 1955.

\bibitem{Bla:J04}
S.~S. Blackman, ``Multiple hypothesis tracking for multiple target tracking,''
  \emph{{IEEE} Trans. Aerosp. Electron. Syst.}, vol.~19, pp. 5--18, Jan. 2004.

\bibitem{KscFreLoe:01}
F.~R. Kschischang, B.~J. Frey, and H.-A. Loeliger, ``Factor graphs and the
  sum-product algorithm,'' \emph{{IEEE} Trans. Inf. Theory}, vol.~47, no.~2,
  pp. 498--519, Feb. 2001.

\bibitem{YedFreWei:05}
J.~Yedidia, W.~Freemand, and Y.~Weiss, ``Constructing free-energy
  approximations and generalized belief propagation algorithms,'' \emph{{IEEE}
  Trans. Inf. Theory}, vol.~51, no.~7, pp. 2282--2312, July 2005.

\bibitem{KolFri:B09}
D.~Koller and N.~Friedman, \emph{{Probabilistic Graphical Models: Principles
  and Techniques}}.\hskip 1em plus 0.5em minus 0.4em\relax Cambridge, MA: MIT
  Press, 2009.

\bibitem{AruMasGorCla:02}
M.~S. Arulampalam, S.~Maskell, N.~Gordon, and T.~Clapp, ``{A tutorial on
  particle filters for online nonlinear/non-Gaussian {B}ayesian tracking},''
  \emph{{IEEE} Trans. Signal Process.}, vol.~50, no.~2, pp. 174--188, Feb.
  2002.

\bibitem{GorMonSca:05}
M.~Gori, G.~Monfardini, and F.~Scarselli, ``A new model for learning in graph
  domains,'' in \emph{Proc. INNS/IEEE IJCNN-05}, vol.~2, Aug. 2005, pp.
  729--734.

\bibitem{GilSchRilVinDah:17}
J.~Gilmer, S.~S. Schoenholz, P.~F. Riley, O.~Vinyals, and G.~E. Dahl, ``Neural
  message passing for quantum chemistry,'' in \emph{Proc. ICML-17}, Aug. 2017,
  pp. 1263--1272.

\bibitem{YooLiaXioZhaFetUrtZemPit:19}
K.~Yoon, R.~Liao, Y.~Xiong, L.~Zhang, E.~Fetaya, R.~Urtasun, R.~Zemel, and
  X.~Pitkow, ``Inference in probabilistic graphical models by graph neural
  networks,'' in \emph{Proc. IEEE Asilomar-19}, 2019, pp. 868--875.

\bibitem{SatWel:21}
V.~G. Satorras and M.~Welling, ``Neural enhanced belief propagation on factor
  graphs,'' in \emph{Proc. AISTATS-21}, Apr. 2021, pp. 685--693.

\bibitem{LiaMey:21}
M.~Liang and F.~Meyer, ``Neural enhanced belief propagation for cooperative
  localization,'' in \emph{Proc. IEEE SSP-21}, Jul. 2021, pp. 326--330.

\bibitem{GraBauReu:J17}
K.~Granstr\"om, M.~Baum, and S.~Reuter, ``{Extended object tracking:
  Introduction, overview and applications},'' \emph{J. Adv. Inf. Fusion},
  vol.~12, no.~2, pp. 139--174, Dec. 2017.

\bibitem{GraFatSve:J19}
K.~{Granstr\"om}, M.~{Fatemi}, and L.~{Svensson}, ``{Poisson multi-Bernoulli
  mixture conjugate prior for multiple extended target filtering},''
  \emph{{IEEE} Trans. Aerosp. Electron. Syst.}, vol.~56, no.~1, pp. 208--225,
  Feb. 2020.

\bibitem{LiaMey:C22}
M.~Liang and F.~Meyer, ``Neural enhanced belief propagation for data
  association in multiobject tracking,'' in \emph{Proc. FUSION-22}, Jul. 2022.

\bibitem{Loe:04}
H.-A. Loeliger, ``An introduction to factor graphs,'' \emph{{IEEE} Signal
  Process. Mag.}, vol.~21, no.~1, pp. 28--41, 2004.

\bibitem{Wym:07}
{H. Wymeersch}, \emph{{Iterative Receiver Design}}.\hskip 1em plus 0.5em minus
  0.4em\relax {Cambridge University Press}, 2007.

\bibitem{WaiJor:03}
M.~J. Wainwright and M.~I. Jordan, ``Graphical models, exponential families,
  and variational inference,'' University of California, Berkeley, Tech. Rep.
  TR-649, Sep. 2003.

\bibitem{BaySahSha:J08}
M.~Bayati, D.~Shah, and M.~Sharma, ``Max-product for maximum weight matching:
  Convergence, correctness, and {LP} duality,'' \emph{{IEEE} Trans. Inf.
  Theory}, vol.~54, no.~3, pp. 1241--1251, 2008.

\bibitem{WilLau:10}
J.~L. Williams and R.~A. Lau, ``Data association by loopy belief propagation,''
  in \emph{Proc. FUSION-10}, 2010, pp. 1--8.

\bibitem{Poo:B94}
H.~V. Poor, \emph{An Introduction to Signal Detection and Estimation},
  2nd~ed.\hskip 1em plus 0.5em minus 0.4em\relax New York: Springer-Verlag,
  1994.

\bibitem{WilLau:J14}
J.~L. Williams and R.~Lau, ``Approximate evaluation of marginal association
  probabilities with belief propagation,'' \emph{{IEEE} Trans. Aerosp.
  Electron. Syst.}, vol.~50, no.~4, pp. 2942--2959, Oct. 2014.

\bibitem{IofSze:15}
S.~Ioffe and C.~Szegedy, ``Batch normalization: Accelerating deep network
  training by reducing internal covariate shift,'' in \emph{Proc. ICML-15},
  Jul. 2015, pp. 448--456.

\bibitem{GuoPleSunWei:17}
C.~Guo, G.~Pleiss, Y.~Sun, and K.~Q. Weinberger, ``On calibration of modern
  neural networks,'' in \emph{Proc. ICML-17}, Aug. 2017, pp. 1321--1330.

\bibitem{Bishop:B06}
C.~M. Bishop, \emph{Pattern Recognition and Machine Learning (Information
  Science and Statistics)}.\hskip 1em plus 0.5em minus 0.4em\relax Berlin,
  Heidelberg: Springer-Verlag, 2006.

\bibitem{OksCamKalAkb:20}
K.~Oksuz, B.~C. Cam, S.~Kalkan, and E.~Akbas, ``Imbalance problems in object
  detection: A review,'' \emph{{IEEE} Trans. Pattern Anal. Mach. Intell.},
  vol.~43, no.~10, pp. 3388--3415, Mar. 2020.

\bibitem{CaeBanLanVorLioXuKriPanBalBei:20}
H.~Caesar, V.~Bankiti, A.~H. Lang, S.~Vora, V.~E. Liong, Q.~Xu, A.~Krishnan,
  Y.~Pan, G.~Baldan, and O.~Beijbom, ``{nuScenes}: A multimodal dataset for
  autonomous driving,'' in \emph{Proc. CVPR-20}, Jun. 2020, pp.
  11\,621--11\,631.

\bibitem{NeuVan:06}
A.~Neubeck and L.~Van~Gool, ``Efficient non-maximum suppression,'' in
  \emph{Proc. ICPR-06}, vol.~3, Aug. 2006, pp. 850--855.

\bibitem{ShaKirLi:B02}
Y.~Bar-Shalom, T.~Kirubarajan, and X.-R. Li, \emph{{Estimation with
  Applications to Tracking and Navigation}}.\hskip 1em plus 0.5em minus
  0.4em\relax New York, NY: Wiley, 2002.

\bibitem{BerSti:08}
K.~Bernardin and R.~Stiefelhagen, ``Evaluating multiple object tracking
  performance: the {CLEAR} {MOT} metrics,'' \emph{EURASIP J. Image Video
  Proc.}, vol. 2008, pp. 1--10, May 2008.

\bibitem{WuNev:06}
B.~Wu and R.~Nevatia, ``Tracking of multiple, partially occluded humans based
  on static body part detection,'' in \emph{Proc. of CVPR-06}, vol.~1, Jun.
  2006, pp. 951--958.

\bibitem{LecBosDenHenHowHubJac:89}
Y.~LeCun, B.~Boser, J.~Denker, D.~Henderson, R.~Howard, W.~Hubbard, and
  L.~Jackel, ``Handwritten digit recognition with a back-propagation network,''
  in \emph{Proc. NeurIPS-89}, vol.~2, Nov. 1989, pp. 396--404.

\bibitem{KinBa:14}
D.~P. Kingma and J.~Ba, ``Adam: A method for stochastic optimization,''
  \emph{arXiv preprint arXiv:1412.6980}, 2014.

\bibitem{BenSchZel:21}
N.~Benbarka, J.~Schr{\"o}der, and A.~Zell, ``Score refinement for
  confidence-based {3D} multi-object tracking,'' in \emph{Proc. IROS-21}, Sep.
  2021, pp. 8083--8090.

\bibitem{RahGarSve:17}
A.~S. Rahmathullah, {\'A}.~F. Garc{\'\i}a-Fern{\'a}ndez, and L.~Svensson,
  ``Generalized optimal sub-pattern assignment metric,'' in \emph{Proc.
  FUSION-17}, Jul. 2017, pp. 1--8.

\bibitem{LeiMeyHlaWitTufWin:J19}
E.~Leitinger, F.~Meyer, F.~Hlawatsch, K.~Witrisal, F.~Tufvesson, and M.~Z. Win,
  ``{A scalable belief propagation algorithm for radio signal based SLAM},''
  \emph{{IEEE} Trans. Wireless Commun.}, vol.~18, no.~12, Dec. 2019.

\end{thebibliography}

\end{document}